\documentclass{article}

\usepackage[table]{xcolor}
\usepackage{microtype}
\usepackage{graphicx}
\usepackage{subfigure}
\usepackage{booktabs} 

\usepackage{hyperref}

\usepackage[accepted]{icml2024}

\usepackage{amsmath}
\usepackage{amssymb}
\usepackage{mathtools}
\usepackage{amsthm}

\def\cD{{\mathcal{D}}}
\def\cA{{\mathcal{A}}}

\def\cL{{\mathcal{L}}}

\def\ha{{\hat{a}}}
\usepackage[capitalize,noabbrev]{cleveref}

\theoremstyle{plain}
\newtheorem{theorem}{Theorem}[section]

\theoremstyle{definition}

\newtheorem{assumption}[theorem]{Assumption}
\theoremstyle{remark}

\usepackage[textsize=tiny]{todonotes}

\icmltitlerunning{SAM-E: Leveraging Visual Foundation Model with Sequence Imitation for Embodied Manipulation}

\begin{document}

\twocolumn[
\icmltitle{SAM-E: Leveraging Visual Foundation Model with Sequence Imitation \\
            for Embodied Manipulation}



\icmlsetsymbol{equal}{*}

\begin{icmlauthorlist}
\icmlauthor{Junjie Zhang}{ts,comp}
\icmlauthor{Chenjia Bai}{comp}
\icmlauthor{Haoran He}{hkust,comp}
\icmlauthor{Wenke Xia}{rm,comp}
\icmlauthor{Zhigang Wang}{comp}
\icmlauthor{Bin Zhao}{comp}
\icmlauthor{Xiu Li}{ts}
\icmlauthor{Xuelong Li}{comp,telecom}
\end{icmlauthorlist}

\icmlaffiliation{ts}{Tsinghua Shenzhen International Graduate School, Tsinghua University}
\icmlaffiliation{comp}{Shanghai Artificial Intelligence Laboratory}
\icmlaffiliation{hkust}{Hong Kong University of Science and Technology}
\icmlaffiliation{rm}{Renmin University of China}
\icmlaffiliation{telecom}{Institute of Artificial Intelligence (TeleAI), China Telecom, P. R. China}

\icmlcorrespondingauthor{Chenjia Bai}{baichenjia@pjlab.org.cn}

\icmlcorrespondingauthor{Xuelong Li}{xuelong\_li@ieee.org}

\icmlkeywords{Machine Learning, ICML}

\vskip 0.3in
]



\printAffiliationsAndNotice{}  

\begin{abstract}
Acquiring a multi-task imitation policy in 3D manipulation poses challenges in terms of scene understanding and action prediction. Current methods employ both 3D representation and multi-view 2D representation to predict the poses of the robot’s end-effector. However, they still require a considerable amount of high-quality robot trajectories, and suffer from limited generalization in unseen tasks and inefficient execution in long-horizon reasoning. In this paper, we propose \emph{SAM-E}, a novel architecture for robot manipulation by leveraging a vision-foundation model for generalizable scene understanding and sequence imitation for long-term action reasoning. Specifically, we adopt Segment Anything (SAM) pre-trained on a huge number of images and promptable masks as the foundation model for extracting task-relevant features, and employ parameter-efficient fine-tuning on robot data for a better understanding of embodied scenarios. To address long-horizon reasoning, we develop a novel multi-channel heatmap that enables the prediction of the action sequence in a single pass, notably enhancing execution efficiency. Experimental results from various instruction-following tasks demonstrate that SAM-E achieves superior performance with higher execution efficiency compared to the baselines, and also significantly improves generalization in few-shot adaptation to new tasks. 
\vskip -0.2in
\vskip -0.1in
\end{abstract}


\section{Introduction}
Robot manipulation has made significant progress, benefiting from embodied datasets \cite{BridgeData,Open-X,Rh20t}, Imitation Learning (IL) \cite{VIMA,Gato} or Reinforcement Learning (RL) algorithms \cite{robopianist,TD-MPC,shi2023robust,bai2024pessimistic}, and advanced transformer \cite{Qtransformer,ALOHA} or diffusion-based networks \cite{chaineddiffuser,MTdiff,he2024large}. To perform a wide range of complex manipulation tasks in the 3D physical world, it is crucial to understand the 3D scene structure that encompasses object positions, orientations, shapes, occlusions, and the relationships between objects and the environment \cite{survey}. Various methods utilize 3D representations such as voxel patches \cite{qattention,peract}, point clouds \cite{PolarNet,SGR}
to provide 3D localizations for predicting the end-effector poses. However, learning a 3D representation can be computationally expensive. For instance, the voxel-based method \cite{peract} achieves state-of-the-art performance while suffering from cubic scaling of the number of voxels with the resolution, making it prohibitive for larger datasets.

To tackle these challenges, recent studies have investigated feature extraction from single-view images and information aggregation using multi-view transformers \cite{hiveFormer}, which provide enhanced efficiency as the scaling of image patches aligns with the input resolution. For example, recently proposed RVT \cite{RVT} achieves 36 times faster training speeds and better performance than voxel-based approaches. However, learning a multi-view policy still requires a considerable amount of high-quality robot trajectories for imitation, and the resulting policy exhibits limited generalization capabilities for unseen tasks and low execution efficiency in long-horizon reasoning. Motivated by recent research on visual foundation models that leverage web-scale datasets and demonstrate robust zero-shot and few-shot generalization \cite{clip,Blip,diffusion,soda}, we delve further into the multi-view architecture to enhance the generalization capabilities and execution efficiency of 3D manipulation policies in language-following tasks. 

In this paper, we present a novel architecture for robot manipulation that leverages a vision-foundation model for image understanding and sequence imitation for long-horizon reasoning. We name our method \textbf{SAM-E}, as we utilize the Segment Anything Model (\textbf{SAM}) \cite{sam} as the foundation model for \textbf{E}mbodied manipulation. SAM is a prompt-conditioned image segmentation model trained on a large dataset of images and masks. Utilizing SAM as the foundational perception model benefits the scene understanding and generalization in various manipulation scenarios. Moreover, the prompt-conditioned SAM encoder is suitable for language-instructed manipulation by extracting task-relevant visual features according to the task descriptions. Further, we conduct parameter-efficient finetuning for SAM with robot data to enhance the understanding of embodied scenarios. With prompt-guided features, we employ multi-view attention to integrate the view-wise representations with coordinate information for action prediction. 

To improve the efficiency of long-horizon action prediction, we propose a novel prediction head that generates multi-channel pose heatmaps for an action sequence. Subsequently, the heatmaps from different views are back-projected into 3D space to generate scores for a discretized set of 3D points, ultimately determining the 3D positions and rotations of actions. During inference, the action sequence can be predicted in a single pass and executed sequentially, resulting in a notable improvement in execution efficiency compared to previous step-by-step prediction methods. We conduct experiments on various 3D instruction-following tasks from RLBench, consisting of 18 tasks with 249 variations \cite{rlbench}. The results demonstrate that SAM-E achieves superior performance and higher reasoning efficiency compared to baseline methods. Moreover, the visual foundation model greatly enhances the generalization ability of the learned policy in adapting to new tasks with few-shot demonstrations.

\section{Preliminaries}
\label{sec:Preliminary}

\textbf{LC-POMDP.} The problem of language-conditioned robot manipulation can be modeled as a Language-Conditioned Partial Observable Markov Decision Process~(LC-POMDP) formulated as an augmented POMDP~$\mathcal{M}:=(\mathcal{S},\mathcal{O},\mathcal{A},\mathcal{P},\rho_0,\mathcal{L},f,T)$, where $\mathcal{S}$ and $\mathcal{A}$ denote state space and action space separately, $O$ denotes the space of observations, $\mathcal{P}(s|s,a):~\mathcal{S}\times\mathcal{A}\rightarrow\mathcal{S}$ denotes the transition probability or the environment dynamics, 
$\rho_0$ represents the initial state distribution, $\mathcal{L}$ denotes the set of all language instructions, $f(o|s): \mathcal{S}\rightarrow\mathcal{O}$ is the observation function, 
and $T$ represents the episode horizon. We adopt imitation learning without considering the reward function used for RL. For each episode, the robot is given a language instruction $l\in\mathcal{L}$ representing the goal of the current task. At each time step $t$, the robot is required to take action according to a policy $\pi(a_t|o_t,l)$ given the observation $o_t$. 
Since we focus on 3D manipulation, the observation $o_t$ contains multi-view images from cameras at different perspectives.

\textbf{Imitation Learning.}~To address the language-conditioned manipulation tasks, imitation learning~\cite{RVT,RoboFlamingo} allows the agent to mimic a set of expert demonstrations denoted as $\cD:=\{(\tau, l)_{i}\}_{i=0}^{|\cD|}$, where $\tau:=(o_0,a_0,\dots,o_{T-1},a_{T-1},o_T)$ is the expert trajectory, and $l$ represents the language instruction. A common imitation learning objective for the policy $\pi_{\theta}$ is to maximize the likelihood of action conditioned on the language and current state, Formally, the loss function is
\begin{equation}
\label{eq:imitation}
\cL(\theta):=-\mathbb{E}_{(\tau,l)\sim \cD}\left[\sum_{t=0}^{T-1}\log\pi_{\theta}(a_t|o_t,l)\right].
\end{equation}

\textbf{Key-frame Extraction.} To improve the utilization of expert demonstrations, we align with the consensus in 3D manipulation algorithms \cite{q-attention,CoF-q-attention,peract,RVT} by incorporating key-frame extraction for selecting key-frame actions. The key-frame extraction involves a Boolean function $K:\mathbb{R}^{|\cA|}\to\{0,1\}$, which determines whether an action should be identified as a key-frame. For each demonstration $\tau$, a sequence of key-frame actions $\{k_1,k_2,\textellipsis,k_m\}$ is generated by the function $K$ following two simple conditions: (\romannumeral1) the joint-velocities are near zero (occurs when entering pre-grasp poses or a new phase of task), and (\romannumeral2) gripper state has changed (occurs when the object is grasped or released). Based on the function $K$, the imitation objective in Eq.~\eqref{eq:imitation} becomes predicting the `next key-frame action' in the demonstration. In the following, we slightly abuse $a_t$ to represent the next key-frame action of $s_t$ since we adopt the same key-frame extraction process to SAM-E and baselines.

\section{Method}
\label{sec:Method}

\begin{figure*}[ht]
\begin{center}
\centerline{\includegraphics[width=\textwidth]{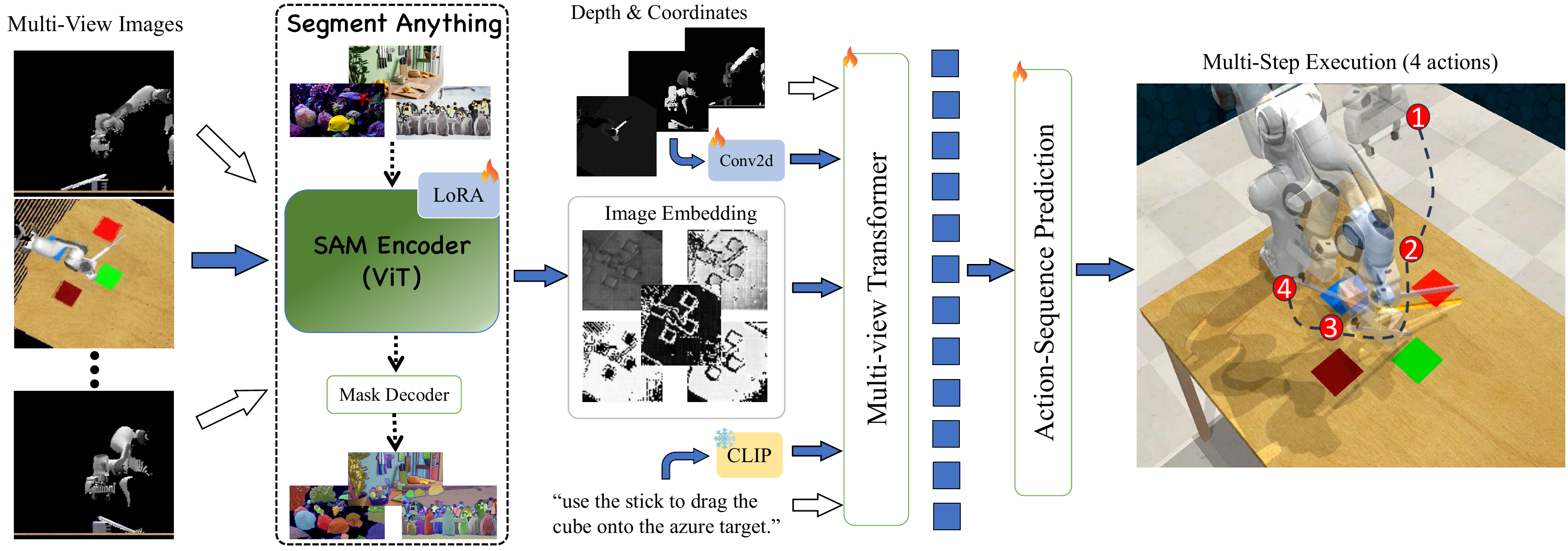}}
\caption{\textbf{Overview of SAM-E.} (\romannumeral1) The SAM encoder provides promptable visual embedding of single-view observations after fine-tuning on embodied scenarios with parameter-efficient LoRA. (\romannumeral2) Multi-view transformer achieves cross-view information integration and vision-language alignment. (\romannumeral3) The coherent action sequence is predicted via temporal imitation for efficient multi-step execution.}
\label{fig:framework}
\end{center}
\vskip -0.2in
\end{figure*}

The proposed \textbf{SAM-E} is a multi-view imitation framework that leverages the pre-trained visual foundation model and action-sequence imitation for multi-task 3D manipulation.
The key idea of SAM-E contains two perspectives: (\romannumeral1) leveraging the visual foundation model SAM with the prompt-driven architecture and its strong generalization ability to handle the language-prompt(instructed) tasks in embodied scenarios; (\romannumeral2) utilizing the temporal smooth properties of actions to perform sequence modeling of actions to enhance coherent planning and execution efficiency. We introduce the visual foundation model for embodied perception in \S\ref{sec:3-1} and the multi-view architecture in \S\ref{sec:3-2}. Then we give the motivation of sequence imitation in \S\ref{sec:3-3} and the multi-channel prediction architecture in \S\ref{sec:3-4}. 

We illustrate the architecture of SAM-E in Figure~\ref{fig:framework}. Overall, we adopt the SAM encoder \cite{sam} to generate prompt-guided and object-oriented representations, and fine-tune the encoder with embodied data and Low-Rank Adaptation (LoRA) \cite{lora} technique for manipulation scenarios, which results in a minimal increase in computation requirement. Then, a multi-view transformer is used to integrate cross-view visual information combined with coordinate information and language instruction for multi-view correspondence and vision-language alignment. To address long-horizon action prediction, SAM-E predicts a coherent action sequence in a single pass with a novel action-sequence policy head.

\subsection{Perception Foundation and LoRA Finetune}
\label{sec:3-1}

\textbf{SAM for Promptable Perception.} SAM \cite{sam} comprises a powerful image encoder and lightweight mask decoder, structured as a prompt-driven architecture designed for real-world image segmentation. Aiming at achieving promptable segmentation and effective ambiguity awareness, the image encoder of SAM is trained with flexible prompts from the downstream mask decoder. Consequently, after diverse segmentation task training, the SAM encoder is capable of extracting powerful object-centering image embedding rich in semantic information. This also enables SAM to handle unknown prompts arising from various segmentation requirements in robot interactions, including complex object-associated scenarios.

In 3D manipulation, the scene perception is expected to be object-oriented and adaptable, accommodating a range of intentions and shifting focus as tasks progress. For instance, given the task instruction of `\emph{place the apple in the basket}', the agent should first find and focus on the apple to pick it up, followed by finding the basket to place. The perception module should be capable of flexible object-centered attention based on task instructions and allow attention adjustment to other objects as the task progresses~(See \S\ref{app:visualization} for an example). From this point, the SAM encoder is suitable as a perception foundation model for language-instructed manipulation with rich task variations. The SAM encoder is a Vision Transformer (ViT) \cite{VIT} pre-trained with MAE \cite{MAE}, which processes RGB images into $C\times H \times W$ image embedding. In practice, we utilize the ViT-B architecture for the image encoder to showcase the advantages of pre-trained segmentation representations with a low computational cost in manipulation tasks. The image encoder contains 12 layers of transformer blocks and outputs the image embedding of the visual inputs. The proposed SAM-E leverages the SAM encoder as the foundation to generate prompt-guided and object-oriented representations from visual observations, which is essential for language-instructed manipulation.

\textbf{LoRA with Embodied Data.} To effectively adapt the SAM encoder to embodied scenarios at an affordable computing cost, we employ LoRA to finetune the encoder during the policy training. As indicated in LoRA, we freeze the parameters in the image encoder and add a trainable low-rank bypass to each of the transformer encoder blocks as:
\begin{equation}
W_0 + \Delta W = W_0 + BA,
\end{equation}

where $W_0 \in \mathbb{R}^{d \times k}$ is the pre-trained weight matrix frozen during training, $B \in \mathbb{R}^{d \times r}$ and $A \in \mathbb{R}^{r \times k}$ are trainable matrix, and rank $r \ll min(d,k)$. $\Delta W = BA$ represents the accumulated gradient update during adaptation with $A$ initiated by Gaussian initialization and $B$ initiated with zero. We set the rank $r$ to 4 by default. In practice, we apply LoRA to the self-attention modules with query and value projection layers:
\begin{equation}
    {\rm Attention}(Q,K,V) = {\rm Softmax}(\frac{QK^T}{\sqrt{d_k}})V,
\end{equation}
\begin{equation}
    Q=\hat{W_q}X=W_qX+B_qA_qX,
\end{equation}
\begin{equation}
    K=W_kX,
\end{equation}
\begin{equation}
    V=\hat{W_v}X=W_vX+B_vA_vX,
\end{equation}
where $W_q$, $W_k$ and $W_v$ are frozen projection weights inherited from SAM encoder, and $A_q$, $B_q$, $A_v$, and $B_v$ are trainable LoRA parameters.

\begin{figure}[t]
\vskip -0.02in
\begin{center}
\centerline{\includegraphics[width=0.9\columnwidth]{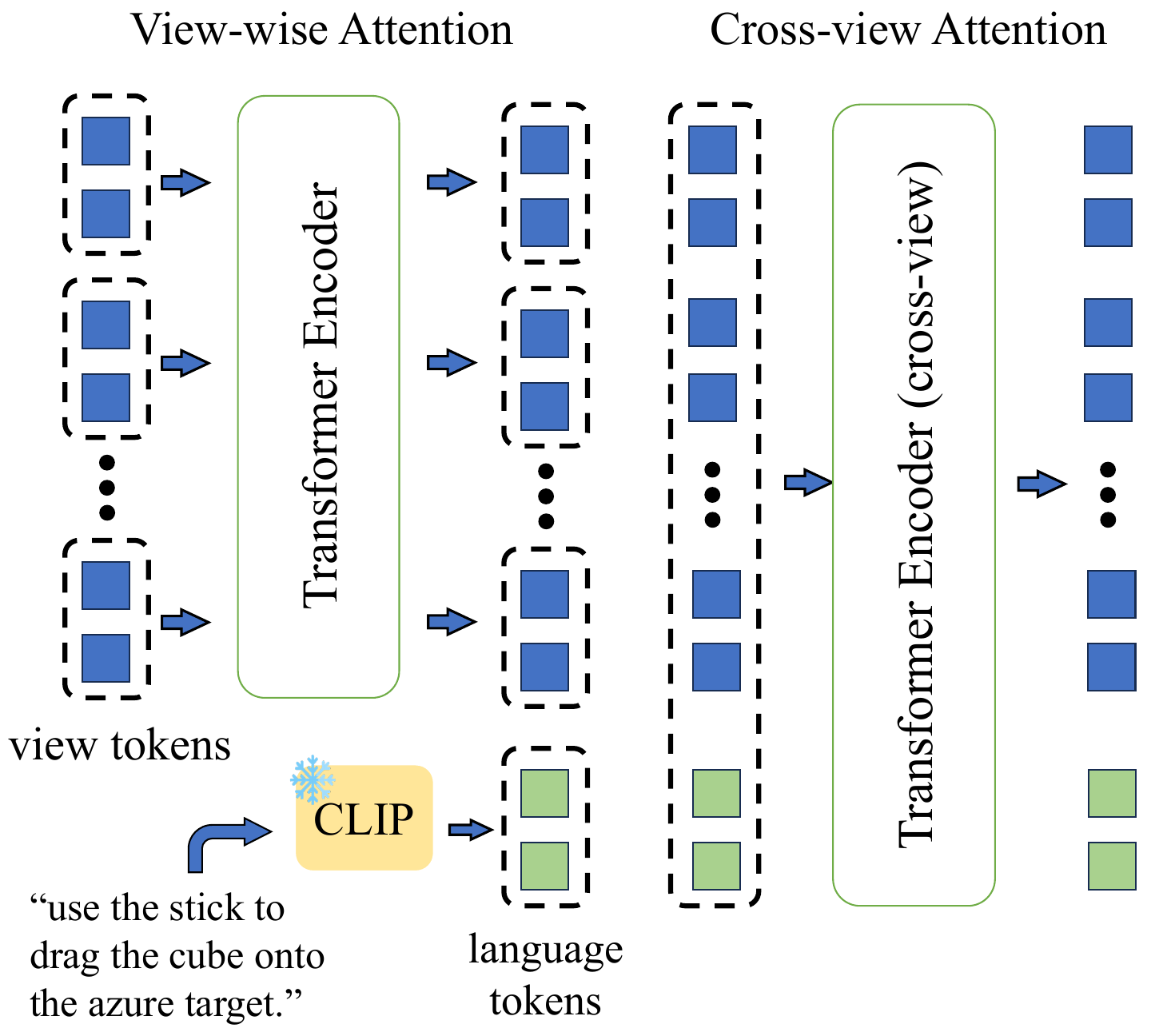}}
\caption{\textbf{Multi-view Transformer} has two stages for view-wise information and cross-view information integration.}
\label{fig:multi-view_transformer}
\end{center}
\vspace{-2em}
\end{figure}

\subsection{Multi-View Transformer} \label{sec:3-2}

After extracting the view-wise representations, we adopt a multi-view transformer to integrate multi-view visual observations, depth information with coordinates, and task-relevant language instructions using an attention mechanism, enabling a comprehensive fusion of the input in multiple modalities. The architecture is shown in Figure~\ref{fig:multi-view_transformer}. The visual observations are processed into image embedding by the previously mentioned SAM encoder, while depth and coordinate information is processed through a Conv2D layer to obtain 3D spatial features. We concatenate the image embeddings with spatial features in the channel dimension along the patch tokens, resulting in a combined representation that we refer to as `view tokens'. Additionally, we utilize a pre-trained CLIP text encoder to generate language embeddings, from which language tokens are derived. Firstly, view tokens from the same view pass through view-wise attention blocks like ViT to maintain the single-view information. Subsequently, visual tokens across different views and the language tokens are attended to cross-view attention blocks, to integrate cross-view scene information with language instructions. The visual tokens, now enriched with cross-view information and language information are used as input for the action-sequence prediction.

\subsection{Motivation for Action-Sequence Modeling}
\label{sec:3-3}

\begin{figure}[t]
\begin{center}
\centerline{\includegraphics[width=\columnwidth]{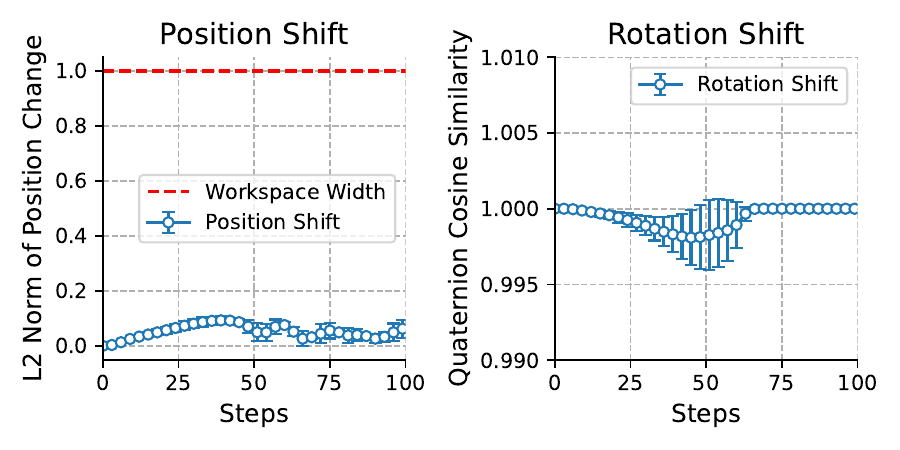}}
\vspace{-0.5em}
\caption{Movement shift in positions and rotations of the end effector in RLBench task \textit{close\_jar}, representing smooth changes of positions and rotations in temporally adjacent steps.}
\label{fig:smooth_action}
\end{center}
\vspace{-1.5em}
\end{figure}

In the next, we aim to provide the intuition of action-sequence modeling, attempting to ground the utility of this technique. We start with an assumption about the temporal smooth properties of actions in the robot manipulation.
\begin{assumption}[Temporal-Smooth Assumption]
\textit{Since the actions of the manipulation task are the desired positions and rotations of the end effector, the optimal action sequences $(a^*_0, a^*_1, a^*_2, \dots, a^*_T)$ are smooth, formulated as:}
\begin{align}
&\| a^*_t - a^*_{t+1} \| < \epsilon,~~0\le t\le T-2,\\
\forall~\tau^*:=(&o_0,a^*_0,o_1,a^*_1,\dots,o_{T-1},a^*_{T-1},o_{T})\sim \mathrm{P}^{\tau}_{\pi^*}(\cdot),\nonumber
\end{align}
\textit{where $\mathrm{P}^{\tau}_{\pi^*}(\cdot)$ denotes the distribution of trajectories derived from the optimal policy $\pi^*$.} 
\label{assum:1}
\end{assumption}
Intuitively, the assumption holds in most embodied manipulation tasks if the actions are the positions and rotations of the end effector. For example, in the common Pick-and-Place tasks, the optimal action sequences are a sequence of points in Euclidean space, which leads the end effector to approach the object and desired goal. Meanwhile, the gripper will rotate smoothly to align with the gripping points of the object. In Figure~\ref{fig:smooth_action}, we show the movement shift of positions and the Quaternion angle of rotations of the end-effector in a manipulation task \textit{close jar} from RLBench \cite{rlbench}, which further justifies our assumption. We observe that in certain manipulation tasks, end effector rotations undergo relatively rapid changes, particularly with large keyframe intervals, which weaken the assumption of smooth rotation. However, the end effector positions maintain superior smoothness in Euclidean space, which are more crucial for action-sequence modeling in our method.

The typical approach trains the policy $\pi$ to predict the action $a_t$ given the multi-view image $o_t$ and the task instruction $l$, as
\begin{center}
$\pi(l,o_t)\to a_t$.
\end{center}
Such a step-by-step process only focuses on predicting actions of the current situation, which can lead to stagnation and contradictory sequential actions, as observed in experiments. Based on the Assumption~\ref{assum:1}, we can improve the action-prediction process by considering a long-horizon decision process instead of a single action, as
\begin{center}
$\pi^{\rm seq}(l,o_t)\to \{a_t,a_{t+1},\textellipsis,a_{t+h-1}\}$,
\end{center}
where $h$ is the horizon of the action sequence. 

Then we motivate the sequence-prediction procedure based on the assumption. The sequence modeling process tries to predict the optimal action sequence condition on the observation. Intuitively, the learning objective of $\pi^{\rm seq}$ is more difficult compared to that of $\pi^{\rm step}$. However, when we take a closer look at the prediction of action in a sequence~(e.g., $a_{t+k}$), training to predict this action is accompanied by the prediction of former actions~(i.e., $(\ha_{t}, \dots, \ha_{t+k-1})$) and latter actions~(i.e., $(\ha_{t+k+1}, \dots, \ha_{t+h-1})$). Back to the assumption that the optimal action sequences are smooth, we believe that predicting the former and latter actions can provide \emph{implicit prior} and \emph{constraint} in predicting $a_{t+k}$. 
Thus, the smooth properties of action sequences provide an opportunity to perform long-horizon reasoning by predicting the adjacent actions as a whole, thereby reflecting the motion trajectory of the robot's end-effector in completing tasks.
In contrast, the action prediction of the traditional policy is only conditioned on the observation without any `prompt' from the former actions, making the traditional policy inferior to action-sequence modeling in these tasks. Such a technique in 2D manipulation tasks is also called action chunking \cite{roboAgent,ALOHA}, while we give a clear motivation by an empirically justified assumption and extend it to 3D scenarios using multi-channel heatmaps.

\subsection{Architecture for Action-Sequence Prediction}
\label{sec:3-4}

\begin{figure}[t]
\vskip 0.2in
\begin{center}
\centerline{\includegraphics[width=\columnwidth]{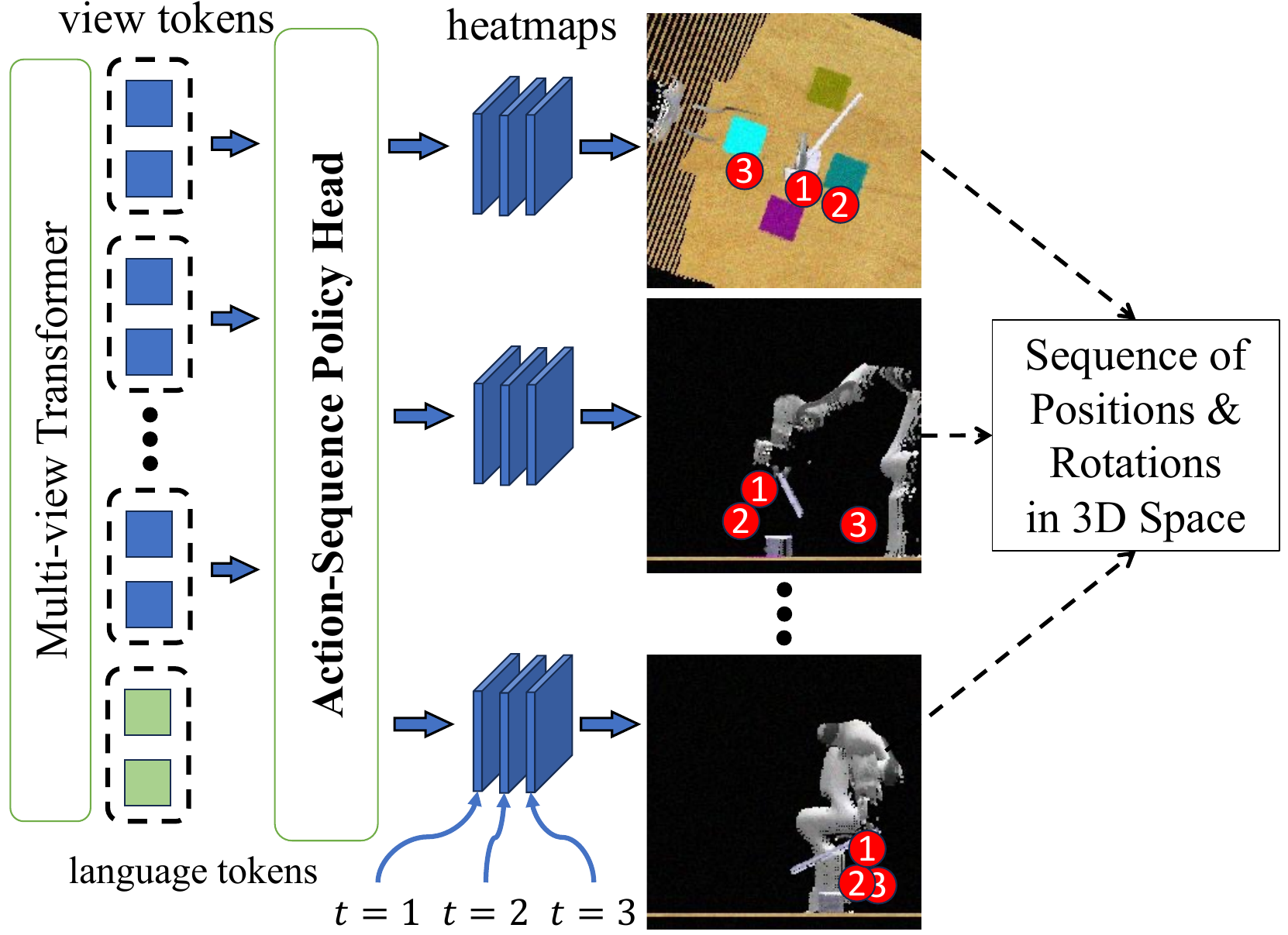}}
\vspace{-0.5em}
\caption{The \textbf{Action-Sequence Policy Head} outputs multi-channel pose heatmaps for a sequence of positions and rotations.}
\label{fig:action_head}
\end{center}
\vspace{-2em}
\end{figure}

We introduce a novel multi-channel policy head for the action-sequence prediction,  as shown in Figure~\ref{fig:action_head}. The policy head takes view tokens from the multi-view transformer (shown in Figure~\ref{fig:multi-view_transformer}) as input, processing view tokens from different views independently, and outputs action sequence prediction in parallel channels within a single view image. 

In 3D manipulation, each action in the sequence comprises an 8-dimensional vector dictating the next movement. This vector includes a 6-DoF target end effector pose (3-DoF for position and 3-DoF for rotation), a binary value indicating the gripper state (open or closed), and another binary value determining whether the collision is permissible for the low-level motion planner. (\romannumeral1) For predicting positions, the policy head generates a heatmap from the view tokens corresponding to each view. These heatmaps represent the desired position distribution from the perspective of each view. Then the heatmaps from different views are back-projected to into 3D space to generate scores for a discretized set of 3D points, determining the 3D positions. For action-sequence prediction, we equip the heatmap with time-dimension channels to learn temporal information from demonstrations, which leads to coherent action prediction in the temporal dimension. (\romannumeral2)
For predicting rotations, we follow previous methods \cite{RVT} to discretize Euler angles into bins of $5^\circ$ resolution and thus turn rotation prediction into classification as the binary of gripper state and collision indicator. We use heatmap as the weight to extract the view-wise features from the view tokens, which provide higher weights near the desired target position within the view image, and then output the action sequence of rotation, gripper state, and collision indicator using a fully connected network. 


\section{Related Works}

\paragraph{Visual Robot Manipulation.} 
Early research in robot manipulation adopts joint states of the robot arm and geometric information of objects in RL or IL frameworks \cite{zeng2017multi,deng2020self,xie2020best,yu2020meta,PromptDT}, assuming the acquisition of pre-perception information and coordinates of objects. In real-world manipulation tasks, visual perception provides more general inputs without additional assumptions \cite{ViGen}. Various methods have adopted visual pretraining models for affordance \cite{goyal2022human,vrb}, representation learning \cite{clip-1,cliport,R3M,VIP,LIV}, and goal generation \cite{VisualGoal,jia2023chain} to facilitate policy learning. Other works incorporate language encoders \cite{xie2023language} and cross-modal transformers \cite{RT-1,hiveFormer} for instruction-following manipulation.	However, these methods learn manipulation policies from top-down 2D images and are limited to pick-and-place primitives \cite{pretrain}. In contrast, by leveraging 3D perception, the robot is able to take into account object orientations, occlusions, and collisions in complex manipulation tasks.	Recent methods utilize 3D representations, such as voxel patches \cite{qattention,peract}, point clouds \cite{PolarNet,SGR,FlowBot3D}, and feature fields \cite{Act3D}, to achieve accurate 3D localizations for action prediction. Another line of research utilizes multi-view images to represent the projections of a 3D environment onto image planes, significantly reducing the computation requirements \cite{liu2023instructionfollowing,MaskedWorld,RVT}. Our method lies in multi-view architectures and leverages pre-trained foundation models to enhance generalization across various visual scenarios and task descriptions. The technique of action chunking is also employed in 2D manipulation \cite{roboAgent,ALOHA}, while we extend it to 3D scenarios using multi-channel heatmaps.

\paragraph{Foundation Models for Embodied Agents.} Large Language Models (LLMs) \cite{touvron2023llama2,GPT4V}, Vision Language Models (VLMs) \cite{liu2023visual,li2023videochat}, and vision foundation models \cite{clip} have demonstrated remarkable capabilities \cite{akyurek2023what} and hold great promise for solving complex embodied tasks.	The chain-of-thought capacity \cite{wei2023chainofthought} of LLMs has been effectively utilized in task planning for embodied agents, including EmbodiedGPT \cite{embodiedgpt}, ReAct \cite{react}, SayCan \cite{ahn2022i}, and DoReMi \cite{guo2023DoReMi}. The commonsense knowledge within LLMs can serve as a world model \cite{zhao2023large,hao2023reasoning,ano2023learning} in text-based environments. Additionally, it can be utilized as a reward designer, as demonstrated by VoxPoser \cite{voxposer}, Text2Reward \cite{Text2Reward}, and Eureka \cite{Eureka}. GenSim \cite{Gensim} and RoboGen \cite{RoboGen} leverage LLMs to generate task curricula and simulation environments to augment robot data. VLMs are commonly employed as foundation models for embodied policies, taking visual observations and language instructions as inputs, and generating language plans \cite{PaLM-E} or tokenized actions \cite{RT-2,GR1} as outputs.	Other approaches utilize VLMs for reward generation \cite{rocamonde2023vision} in RL frameworks and self-reflection for task planning \cite{hu2023look}. RoboFlamingo \cite{RoboFlamingo} is related to our method as it employs OpenFlamingo as a base policy and finetunes this policy using embodied datasets. However, it is limited to 2D manipulation and lacks explicit consideration of 3D geometry, which hinders its capacity to develop highly accurate spatial manipulation skills in robotics.

\paragraph{Segment Anything Model.} 
SAM \cite{sam} is a promptable segmentation model capable of generating masks by receiving various prompts, including points, bounding boxes, and language prompts. Subsequent works have examined the application of SAM for object localization \cite{zhang2023personalize}, tracking \cite{rajivc2023segment,cheng2023segment}, and semantic analysis \cite{mazurowski2023segment}. For embodied agents, SAM-G \cite{samg} is a concurrent work that utilizes point prompts to establish correspondences and employs SAM to generate masked images for the agent. However, SAM-G focuses on extracting the agent-relevant mask for robust visual representations and mitigating the impact of noise (e.g., colors, backgrounds) in 2D manipulation and locomotion tasks. In contrast, our method adopts SAM to enhance 3D manipulation within a multi-view framework and extracts task-relevant features to facilitate generalization across various manipulation scenarios and language instructions. 

\section{Experiments}

In this section, we evaluate SAM-E in RLBench \cite{rlbench}, which is a challenging multi-task 3D manipulation benchmark. To perform a fair comparison to baselines, we use the same settings as the state-of-art method \cite{RVT} by using 18 tasks with 249 variations in experiments. Moreover, we evaluate the generalization ability of SAM-E via few-shot adaptation in 6 new tasks. The Videos are available at: \url{https://sam-embodied.github.io/}.

\subsection{Experiment Setup}
\textbf{Baselines.} We compare SAM-E against off-the-shelf algorithms proved to work on multi-view 3D manipulation, including (\romannumeral1) \textbf{RVT} \cite{RVT}, the state-of-the-art multi-view architecture for 3D manipulation by re-rendering visual observations into orthographic projections of cube views and predicting the next move based on these projections; (\romannumeral2) \textbf{PerAct} \cite{peract}, an action-centric method that encodes RGB-D images into voxel grid patches for 3D representation and predicts the action within the 3D voxel space. (\romannumeral3) We include \textbf{R3M} \cite{R3M}, the visual representation designed for robot manipulation, as an alternative encoder in our architecture. (\romannumeral4) We include two more general visual representations CLIP \cite{clip}, DINO \cite{dino}  in our architecture. (\romannumeral5) We include a variant referred to \textbf{SAM$\rightarrow$RVT} that replaces the SAM encoder with RVT's visual encoder, which is trained from scratch. (\romannumeral6) Since RVT has been shown to significantly outperform other behavior cloning (BC) baselines including \textbf{CNN-BC}, \textbf{ViT-BC} \cite{jang2021bcz}, and \textbf{Coarse-to-Fine BC} \cite{qattention}, we do not include the scores of these methods and we refer to \citet{RVT} for details. (\romannumeral7) Additionally, we compare SAE-E against \textbf{Hiveformer} \cite{hiveFormer} with same tasks evaluated in their paper (we refer to \S\ref{app_hive} for the results).


\textbf{Simulation Environment.} We perform experiments in RLBench \cite{rlbench}, which is simulated by CoppeliaSim \cite{rohmer2013v} to control a Franka Panda robot equipped with a parallel gripper. Visual observations are captured by four RGB-D cameras (left shoulder, right shoulder, front, and wrist) with a resolution of 128 $\times$ 128, and target gripper pose is achieved by a sample-based motion planner. In this elaborated simulator, the agent is tested to complete the task within a limited number of timesteps, which is 25 in experiments. The tasks include picking and placing items, executing staged moves for tool usage, and comprehending scenes to solve puzzles (see \S\ref{app:task} for more detailed descriptions of the tasks). The algorithms are evaluated in a multi-task and multi-modal setting, characterized by a high degree of variation, which necessitates the agent to demonstrate scene understanding, instruction comprehension, and precise action prediction.

\textbf{Training Datasets.} We utilize the same training datasets as RVT and PerAct, comprising 100 expert demonstrations per task. Unlike RVT and PerAct, which slice demonstration episodes into keyframe transitions with empirically crucial duplication for important transitions, we seamlessly decompose demonstrations into multiple sub-episodes of keyframes to facilitate action-sequence prediction.
We train SAM-E for 60K steps and choose the last model for evaluation, which is the same as RVT. We use cosine learning rate decay after 2K steps warm-start (see \S\ref{app:implement} for more details).

\begin{figure}[t]
\begin{center}
\centerline{\includegraphics[width=0.9\columnwidth]{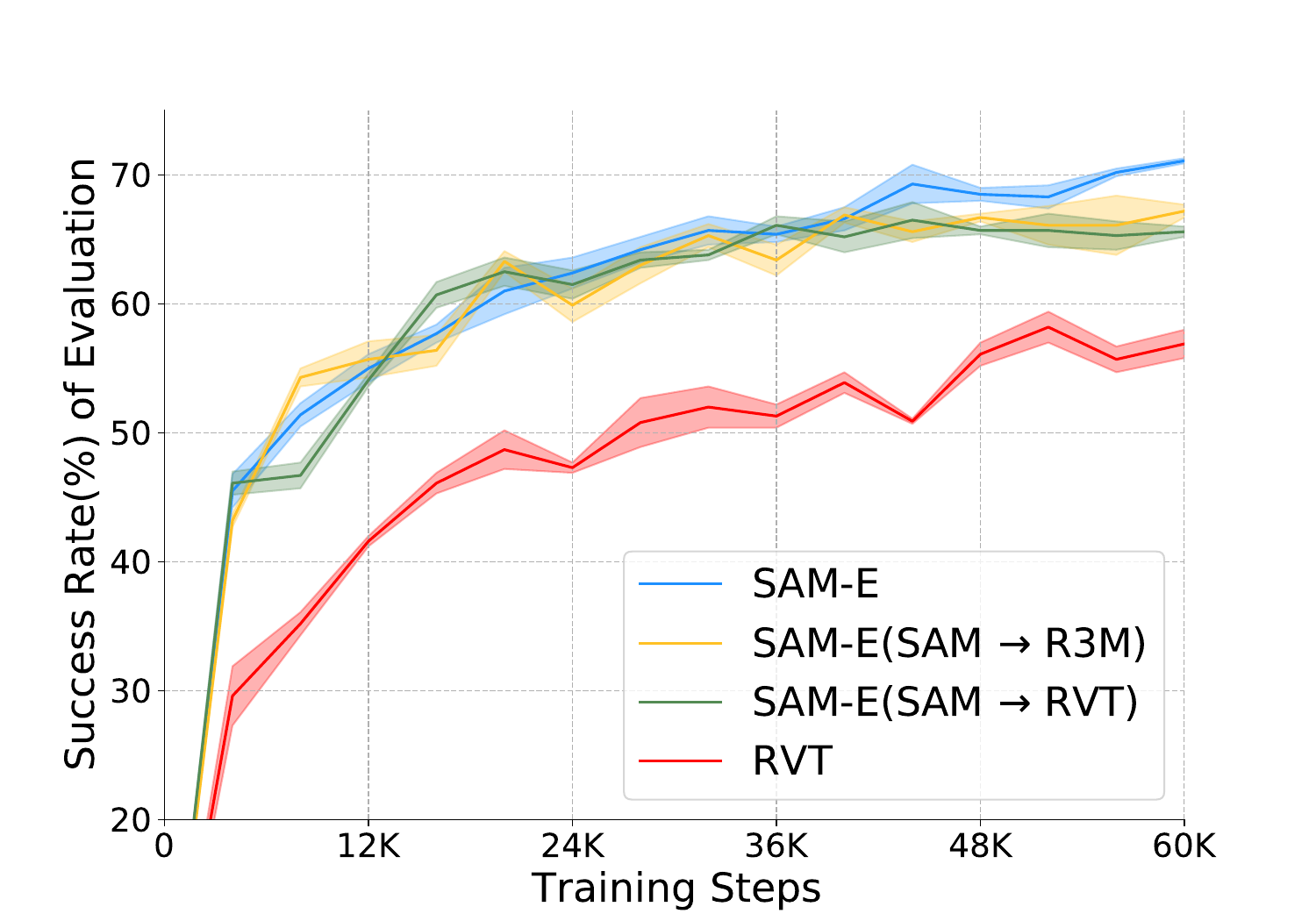}}
\vspace{-0.5em}
\caption{The comparison of training curves from 5 seeds with $\pm$1 std. We observe that SAM-E achieves a higher success rate than R3M and non-pre-trained baselines. Meanwhile, SAM and its variations achieve a better training efficiency compared to RVT, benefiting from action sequence imitation. The training curve of RVT is from our reproduction by running the official code.}
\label{fig:training_curve}
\end{center}
\vspace{-2em}
\end{figure}

\subsection{Main Experiments}

\begin{figure}[t]
\begin{center}
\centerline{\includegraphics[width=0.8\columnwidth]{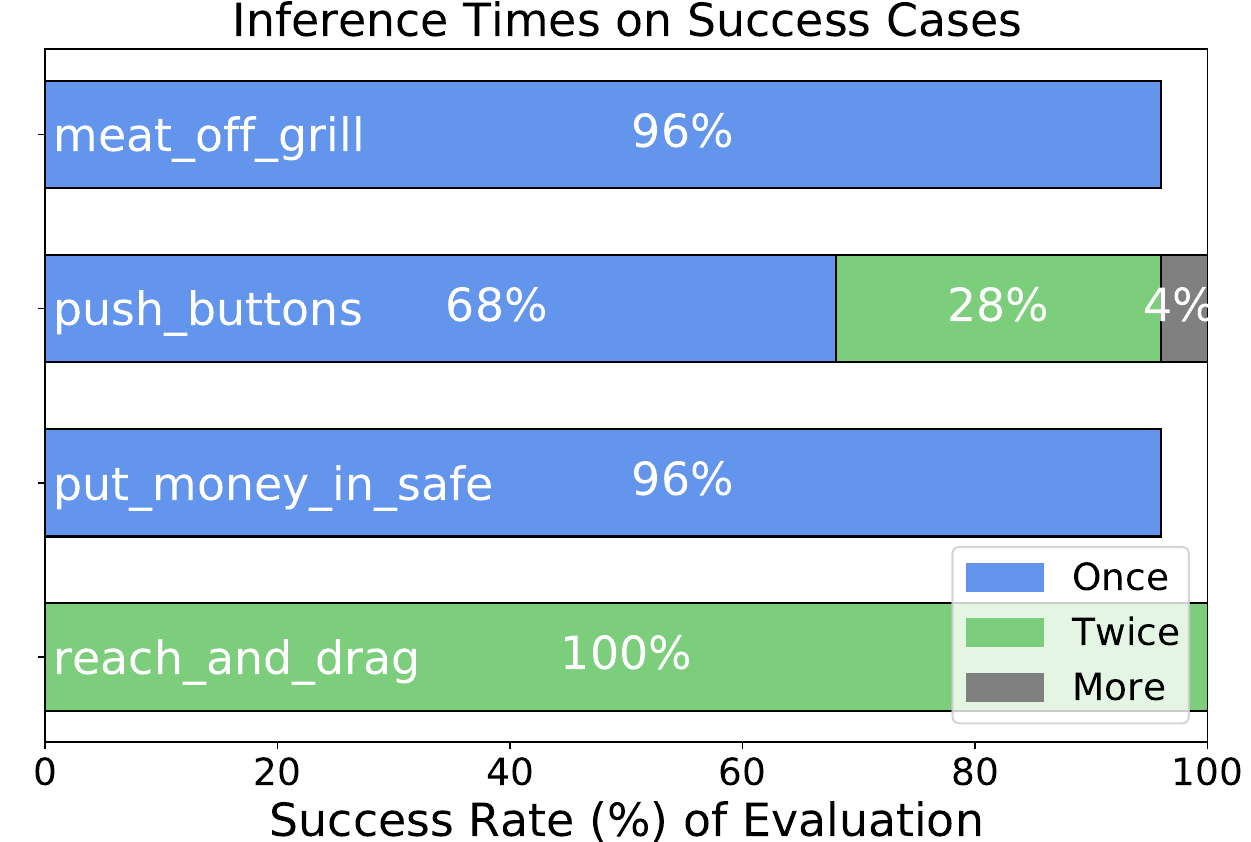}}
\vspace{-0.5em}
\caption{An illustration of the execution efficiency in several tasks. SAM-E completes most tasks in merely once or twice inferences in all success cases. We refer to \S\ref{app:infer} for more examples.}
\label{fig:one_glance}
\end{center}
\vspace{-2em}
\end{figure}



\begin{table}[]
\centering
\caption{A comparison of trainable parameters to baselines.}
\vspace{0.2em}
\label{tab:model_parameter}
\resizebox{\columnwidth}{!}{
\begin{tabular}{@{}lccccc@{}}
\toprule
Models  & RVT & SAM-E~(SAM $\to$ RVT) & SAM-E~(SAM $\to$ R3M) & SAM-E  \\
\midrule
Trainable Para.                & 36.3M	& 35.6M & 35.6M & 35.7M	  \\
\bottomrule
\end{tabular}
}
\vspace{-2em}
\end{table}

\begin{table*}[t]
\centering
\caption{\textbf{Multi-task Performances.} SAM-E outperforms state-of-the-art methods in most tasks and on average, with much fewer inference steps in execution. Scores of PerAct and RVT are adopted from \citet{RVT}. Mean and std of 5 evaluations are reported.}
\label{tab:main_experiment_0}
\resizebox{\textwidth}{!}{
\begin{tabular}{@{}lcccccccccc@{}}
\toprule
Models & \begin{tabular}[c]{@{}c@{}}Put in\\ Drawer\end{tabular} & \begin{tabular}[c]{@{}c@{}}Reach \\and Drag\end{tabular} & \begin{tabular}[c]{@{}c@{}}Turn\\ Tap\end{tabular} & \begin{tabular}[c]{@{}c@{}}Slide to\\ Target\end{tabular} & \begin{tabular}[c]{@{}c@{}}Open\\ Drawer\end{tabular} & \begin{tabular}[c]{@{}c@{}}Put in\\ Cupboard\end{tabular} & \begin{tabular}[c]{@{}c@{}}Place in\\ Shape Sorter\end{tabular} & \begin{tabular}[c]{@{}c@{}}Put Money\\ in Safe\end{tabular} & \begin{tabular}[c]{@{}c@{}}Push\\ Buttons\end{tabular} & \begin{tabular}[c]{@{}c@{}}Close\\ Jar\end{tabular}\\ 
\midrule
PerAct & 51.2$\pm$4.7 &	89.6$\pm$4.1 & 88.0$\pm$4.4 & 74.0$\pm$13.0	&88.0$\pm$5.7 & 28.0$\pm$4.4&16.8$\pm$4.7&84.0$\pm$3.6&92.8$\pm$3.0 &55.2$\pm$4.7 \\
RVT & 88.0$\pm$5.7 & 99.2$\pm$1.6 & 93.6$\pm$4.1 & 81.6$\pm$5.4 & 71.2$\pm$6.9 & 49.6$\pm$3.2 & 36.0$\pm$2.5 & 91.2$\pm$3.0 & \textbf{100.0}$\pm$0.0  & 52.0$\pm$2.5 \\
SAM-E~(SAM $\to$ RVT) & 87.2$\pm$5.9	&\textbf{100.0}$\pm$0.0	&\textbf{100.0}$\pm$0.0	&79.2$\pm$6.6	&\textbf{95.2}$\pm$3.3	&59.2$\pm$5.2	&35.2$\pm$4.4	&72.0$\pm$4.0	&98.4$\pm$2.2	&83.2$\pm$5.9\\
SAM-E~(SAM $\to$ R3M) & 83.2$\pm$5.9	&99.2$\pm$1.8	&\textbf{100.0}$\pm$0.0	&88.8$\pm$4.4	&\textbf{95.2}$\pm$3.3	&41.6$\pm$7.3	&31.2$\pm$7.7	&\textbf{95.2}$\pm$3.3	&96.0$\pm$0.0 & 78.4$\pm$2.2\\
SAM-E~(SAM $\to$ CLIP) & 88.8$\pm$3.3	&\textbf{100.0}$\pm$0.0	&\textbf{100.0}$\pm$0.0	&78.4$\pm$13.4	&92.0$\pm$4.0	&40.0$\pm$4.9	&\textbf{42.4}$\pm$6.1	&80.8$\pm$1.8	&\textbf{100.0}$\pm$0.0 & 73.6$\pm$2.2\\
SAM-E~(SAM $\to$ DINO) & 78.4$\pm$4.6	&99.2$\pm$1.8	&99.2$\pm$1.8	&88.0$\pm$4.9	&89.6$\pm$5.4	&52.0$\pm$7.5	&30.4$\pm$9.2	&85.6$\pm$2.2	&\textbf{100.0}$\pm$0.0 & \textbf{89.6}$\pm$3.6\\
SAM-E~(ours) & \textbf{92.0}$\pm$5.7 & \textbf{100.0}$\pm$0.0 & \textbf{100.0}$\pm$0.0 & \textbf{95.2}$\pm$1.8 & \textbf{95.2}$\pm$5.2 & \textbf{64.0}$\pm$2.8 & 34.4$\pm$6.1 & \textbf{95.2}$\pm$3.3 & \textbf{100.0}$\pm$0.0 & 82.4$\pm$3.6 \\
\bottomrule
Models  & \begin{tabular}[c]{@{}c@{}}Stack\\ Blocks\end{tabular} & \begin{tabular}[c]{@{}c@{}}Place\\ Cups\end{tabular} & \begin{tabular}[c]{@{}c@{}}Place Wine\\ at Rack\end{tabular} & \begin{tabular}[c]{@{}c@{}}Screw\\ Bulb\end{tabular} & \begin{tabular}[c]{@{}c@{}}Sweep to\\ Dustpan\end{tabular} & \begin{tabular}[c]{@{}c@{}}Insert \\ Peg\end{tabular} &
\begin{tabular}[c]{@{}c@{}}Meat off\\ Grill\end{tabular} & 
\begin{tabular}[c]{@{}c@{}}Stack\\ Cups\end{tabular} &
\multicolumn{1}{|c}{\begin{tabular}[c]{@{}c@{}}\textbf{On}\\ \textbf{Average}\end{tabular}} &
\begin{tabular}[c]{@{}c@{}}\textbf{Inference}\\ \textbf{Steps(Sum)}\end{tabular}\\
\midrule
PerAct &26.4$\pm$3.2	&2.4$\pm$3.2	&44.8$\pm$7.8	&17.6$\pm$2.0	&52.0$\pm$0.0	&5.6$\pm$4.1	&70.4$\pm$2.0	&2.4$\pm$2.0	&\multicolumn{1}{|c}{49.4} & -\\
RVT  & 28.8$\pm$3.9 & \textbf{4.0}$\pm$2.5 & 91.0$\pm$5.2 & 48.0$\pm$5.7 & 72.0$\pm$0.0 & 11.2$\pm$3.0 & 88.0$\pm$2.5 & \textbf{26.4}$\pm$8.2 & \multicolumn{1}{|c}{62.9} & 6158$\pm$64\\
SAM-E~(SAM $\to$ RVT) &22.4$\pm$3.6	&0.0$\pm$0.0	&92.8$\pm$6.6	&61.6$\pm$9.2	&84$\pm$0.0	&7.2$\pm$5.9	&95.2$\pm$3.3	&3.2$\pm$3.3	&\multicolumn{1}{|c}{65.3$\pm$0.6}	&1190$\pm$19 \\
SAM-E~(SAM $\to$ R3M) &\textbf{32.0}$\pm$2.8	&1.6$\pm$2.2	&92.8$\pm$3.3	&60.0$\pm$2.8	&96.8$\pm$3.3	&5.6$\pm$6.7	&\textbf{97.6}$\pm$2.2	&2.4$\pm$2.2	&\multicolumn{1}{|c}{66.5$\pm$1.0} &1165$\pm$63 \\
SAM-E~(SAM $\to$ CLIP) &22.4$\pm$10.8	&0.0$\pm$0.0	&93.6$\pm$2.2	&59.2$\pm$4.4	&85.6$\pm$2.2	&8.0$\pm$2.8	&96.0$\pm$4.0	&4.8$\pm$3.3	&\multicolumn{1}{|c}{64.8$\pm$0.9} &1192$\pm$17 \\
SAM-E~(SAM $\to$ DINO) &28.8$\pm$7.7	&0.8$\pm$1.8	&93.6$\pm$3.6	&64.0$\pm$9.8	&\textbf{100.0}$\pm$0.0	&11.2$\pm$3.3	&96.0$\pm$2.8	&1.6$\pm$2.2	&\multicolumn{1}{|c}{67.1$\pm$0.4} &1143$\pm$15 \\
SAM-E~(ours)  & 26.4$\pm$4.6 & 0.0$\pm$0.0 & \textbf{94.4}$\pm$4.6 & \textbf{78.4}$\pm$3.6 & \textbf{100.0}$\pm$0.0 & \textbf{18.4}$\pm$4.6 & 95.2$\pm$3.3 & 0.0$\pm$0.0 & \multicolumn{1}{|c}{\cellcolor{blue!20}\textbf{70.6}$\pm$0.7} &\cellcolor{blue!20} \textbf{1130}$\pm$12\\
\bottomrule
\end{tabular}
}
\vspace{-1em}
\end{table*}

\begin{table*}[htbp]
\centering
\caption{\textbf{Few-shot adaptation.} Mean and std of 5 evaluations are reported.}
\label{tab:fewshot_adaptation}
\resizebox{0.85\textwidth}{!}{
\begin{tabular}{@{}lcccccc|c@{}}
\toprule
Models  & \begin{tabular}[c]{@{}c@{}}Meat on\\ Grill\end{tabular} & \begin{tabular}[c]{@{}c@{}}Open\\ Jar\end{tabular} & \begin{tabular}[c]{@{}c@{}}Screw\\ Nail\end{tabular} & \begin{tabular}[c]{@{}c@{}}Toilet \\Seat Done\end{tabular} & \begin{tabular}[c]{@{}c@{}}TV\\ on\end{tabular} & \begin{tabular}[c]{@{}c@{}}Solve\\ Puzzle\end{tabular} & \begin{tabular}[c]{@{}c@{}}\textbf{On}\\ \textbf{Average}\end{tabular} \\
\midrule
RVT~~~~~~(from scratch)                    & \textbf{80.0}$\pm$6.3	&\textbf{36.0}$\pm$4.9	&7.2$\pm$4.4	&99.2$\pm$1.8	&2.4$\pm$3.6	&11.2$\pm$4.4	&39.3$\pm$2.3  \\
SAM-E~(from scratch, SAM $\to$ RVT)        & 60.0$\pm$2.8	&12.0$\pm$0.0	&\textbf{36.0}$\pm$6.9	&96.0$\pm$0.0	&15.2$\pm$3.3	&20.8$\pm$7.7	&40.0$\pm$1.8\\
SAM-E~(from scratch, SAM $\to$ R3M)  & 69.6$\pm$4.6&	16.0$\pm$0.0	&29.6$\pm$6.1&	\textbf{100.0}$\pm$0.0	&12.0$\pm$2.8	&\textbf{22.4}$\pm$2.2&	41.6$\pm$1.4 \\
SAM-E~(from scratch, SAM $\to$ CLIP)  & 64.0$\pm$0.0&	12.0$\pm$0.0	&16.0$\pm$4.0&	\textbf{100.0}$\pm$0.0	&14.7$\pm$6.1	&\textbf{24.0}$\pm$4.0&	38.4$\pm$1.0 \\
SAM-E~(from scratch, SAM $\to$ DINO)  & 53.3$\pm$8.3&	12.0$\pm$4.0	&26.7$\pm$2.3&	\textbf{100.0}$\pm$0.0	&16.0$\pm$4.0	&\textbf{24.0}$\pm$4.0&	38.7$\pm$1.2 \\
SAM-E~(from scratch)                 & 75.2$\pm$4.4	&12.8$\pm$1.8	&28.0$\pm$8.0	&\textbf{100.0}$\pm$0.0	&\textbf{20.8}$\pm$1.8	&17.6$\pm$2.2	&\cellcolor{blue!20}\textbf{42.4}$\pm$1.5 \\
\midrule
RVT~~~~~~(adaptation)                     &68.8$\pm$3.3	&36.0$\pm$0.0	&1.6$\pm$2.2	&\textbf{100.0}$\pm$0.0	&1.6$\pm$2.2	&14.4$\pm$6.7	&37.1$\pm$1.0\\
SAM-E~(adaptation, SAM $\to$ RVT)         & 69.6$\pm$6.1	&39.2$\pm$3.3	&38.4$\pm$4.6	&99.2$\pm$1.8	&17.6$\pm$2.2	&38.4$\pm$3.6	&50.4$\pm$1.1 \\
SAM-E~(adaptation, SAM $\to$ R3M)   &64.8$\pm$5.9	&37.6$\pm$2.2	&28.8$\pm$5.9&	\textbf{100.0}$\pm$0.0	&12.8$\pm$1.8	&37.6$\pm$6.7&	46.9$\pm$2.3\\
SAM-E~(adaptation, SAM $\to$ CLIP)   &78.7$\pm$2.3	&38.7$\pm$4.6	&28.0$\pm$6.9&	\textbf{100.0}$\pm$0.0	&16.0$\pm$0.0	&25.3$\pm$8.3&	47.8$\pm$1.9\\
SAM-E~(adaptation, SAM $\to$ DINO)   &68.0$\pm$4.0	&33.3$\pm$6.1	&50.7$\pm$8.3&	98.7$\pm$2.3	&24.0$\pm$4.0	&22.7$\pm$2.3&	49.6$\pm$1.0\\
SAM-E~(adaptation)                   & \textbf{84.0}$\pm$5.7	&\textbf{56.0}$\pm$7.5	&\textbf{62.4}$\pm$4.6	&\textbf{100.0}$\pm$0	&\textbf{35.2}$\pm$1.8	&\textbf{41.6}$\pm$7.3	&\cellcolor{blue!20}\textbf{63.2}$\pm$1.5 \\
\bottomrule
\end{tabular}
}
\vspace{-1em}
\end{table*}

\textbf{Multi-Task Learning.} We train all methods in 18 tasks and the comparison of success rate is given in Table \ref{tab:main_experiment_0}. SAM-E outperforms PerAct and RVT in 14 out of 18 tasks. SAM-E outperforms PerAct and RVT by an average of \textbf{21.2\%} and \textbf{7.7\%} percentage points in success rate across 18 tasks, marking a relative improvement with \textbf{43.0\%} and \textbf{12.2\%}, while incurring significantly lower model inference costs. Furthermore, it achieves an improvement exceeding 30\% points in several tasks. Eliminating the pre-trained SAM encoder in SAM-E leads to a performance drop but still outperforms RVT, benefiting from the action sequence policy head. Building upon this, the addition of R3M's frozen representation has yielded a marginal performance improvement, however, which is still inferior compared to SAM-E. Similarly, CLIP and DINO representations have mediocre performances compared to SAM-E. Notably, SAM-E has comparable training time and even less trainable parameters compared to RVT, as shown in Table \ref{tab:model_parameter}. Moreover, Figure \ref{fig:training_curve} shows that SAM-E and its variations exhibit higher training efficiency
than RVT, mainly attributed to the action sequence imitation. Further, utilizing SAM as the scalable visual foundation, SAM-E not only achieves the best performance on the current setup, 
but also shows potential for further enhancing its advantages with more embodied data or update steps.

Different from baselines that predict the next keypoint gripper pose at each timestep, SAM-E generates a sequence of actions for long-term planning and sequential execution, thereby considering the task completion from a higher perspective and has much fewer inference steps. According to Table~\ref{tab:main_experiment_0}, SAM-E demonstrates an average execution efficiency of more than \textbf{5X} greater than that of RVT. In tasks such as \textit{meat\_off\_grill}, \textit{push\_buttons}, and \textit{put\_money\_in\_safe} (see \S\ref{app:task} for task descriptions), SAM-E can complete the task after merely \textbf{a glance} at the initial state, as shown in Figure \ref{fig:one_glance}. In contrast, RVT requires, on average, 5.5, 3.8, and 6.0 steps to complete them for its successful cases. For \textit{reach\_and\_drag}, SAM-E completes it all in two inferences while RVT needs to execute 6.4 times on average.

\textbf{Few-Shot Adaptation.} 
We evaluate the generalization ability of SAM-E by adapting the trained model to 6 new tasks from RLBench. We use 10X fewer demonstrations and 15X fewer update steps in policy adaptation than that of the multi-task experiments to show the generalization capability of the SAM-E in few-shot adaptation. The results are shown in Table \ref{tab:fewshot_adaptation}. We initialize the models with weights from their multi-task training for adaptation, and also introduce their random initialization variants for training from scratch. We find RVT struggles with transferring knowledge from previous tasks to new ones during adaptation, often resulting in performance drops compared to training from scratch. In contrast, SAM-E significantly benefits from adaptation compared to starting from scratch. Specifically, SAM-E outperforms RVT by \textbf{3.1\%} points (a \textbf{7.9\%} relative increase) when trained from scratch. However, during adaptation to new tasks, the performance gap widens dramatically, with SAM-E surpassing RVT by \textbf{26.1\%} points, a substantial \textbf{70.4\%} relative improvement. This demonstrates that SAM-E has superior generalization capabilities. 

When training from scratch, \emph{SAM-E (SAM $\to$ R3M)} achieves a slightly better performance than \emph{SAM-E (SAM $\to$ RVT)} that does not have a pre-trained encoder, but results in worse performance in adaptation, which shows R3M has limited few-shot generalization ability. While worse than \emph{SAM-E (SAM $\to$ R3M)} in training from scratch, \emph{SAM-E (SAM $\to$ CLIP)} and \emph{SAM-E (SAM $\to$ DINO)} have better performances in adaptation, showing greater generalization of the representations pre-trained in more general image data. \emph{SAM-E (SAM $\to$ RVT)} also significantly outperforms RVT in adaptation over from scratch, demonstrating the enhanced generalization ability gained from the action-sequence prediction. In terms of adapting to new tasks, SAM-E equipped with a SAM encoder demonstrates significant advantages over the methods mentioned above. This highlights the exceptional capabilities of SAM-E to generalize in novel task descriptions. 

\begin{table}[h]
\vspace{-0.5em}
\centering
\caption{Success rate and parameters amount of the variations}
\vskip 0.05pt
\label{tab:ablation}
\resizebox{\columnwidth}{!}{
\begin{tabular}{@{}lc|cc@{}}
\toprule
Models & Success Rate & Parameters & Trainable Parameters \\
\midrule
SAM-E               &\cellcolor{blue!20} \textbf{70.6}$\pm$0.7 &122.5M	&35.7M	  \\
SAM-E~(SAM $\to$ RVT)         & 65.3$\pm$0.6  & 35.6M  &35.6M     \\
SAM-E~(LoRA, QKV)     & 69.2$\pm$0.9  & 122.6M   &35.8M        \\
SAM-E~(w/o LoRA)     & 67.2$\pm$1.0  &122.4M   &35.6M        \\
SAM-E~(full finetune)  & 65.8$\pm$1.0   &122.4M  &122.4M        \\
\bottomrule
\end{tabular}
}
\end{table}
\subsection{Ablations}
First, we conduct ablation experiments in multi-task experiments to verify the necessity of components in SAM-E. We include (\romannumeral1) \emph{SAM-E (SAM $\to$ RVT)}; (\romannumeral2) \emph{SAM-E (LoRA, QKV)}, which is a variant of LoRA module additionally including $K$ matrix of attention blocks; (\romannumeral3) \emph{SAM-E (w/o LoRA)}, a frozen SAM encoder without LoRA fine-tuning, and (\romannumeral4) \emph{SAM-E (full finetune)}, which performs full-parameter training of the SAM encoder. We give the brief result in Table \ref{tab:ablation}. We find SAM is a crucial visual foundation and a suitable finetune method is required for adaptation to embodied scenarios. Using LoRA to parameter-efficiently finetuning, SAM is better than the variant that trains all parameters, which may lead to failure due to the limited demonstrations. For LoRA, adding the trainable matrix for $Q$ and $V$ is better than all $Q$, $K$, and $V$, which is consistent with previous observations \cite{lora}. (See \S\ref{app:ablation} for the complete results)

Additionally, to illustrate the impact of the action sequence length $h$ (refer to \S\ref{app:horizon} for details), we conduct an ablation study on the action horizon, examining $h$ values of \{1,3,5,7\}. During both the training and evaluation execution of the multi-task experiments, we modify the action horizon $h$ while maintaining consistency in other experimental settings. The outcomes are presented in Table \ref{tab:ablation_horizon} (See \S\ref{app:ablation} for the complete results), showing the average success rate across 18 tasks and the computing time for each model inference on our same device during the model evaluation.  We observe that $h=5$ performs the best on the average success rate, while it may not suitable for certain tasks. We can also find that $h=1$ leads to a drop in performance, which we attribute to the insufficient temporal information to drive SAM foundation training, combined with the lack of empirically crucial duplication for important transitions. Moreover, we can observe that SAM-E's inference time is slightly longer than that of RVT. Nevertheless, SAM-E is even faster in inference considering an action sequence (5 actions) is predicted in 152ms, while RVT requires 5*103ms to predict 5 actions.

\subsection{Real-World Experiment}
To demonstrate the effectiveness of SAM-E in real-world scenarios, we train and test the model in a real-world setup with a Franka Panda robot arm. As shown in Figure \ref{appendix_fig:realworld}, we use two statically mounted RGB-D cameras in a third-person view at the left front and right front to capture the multi-view observation. We calibrate the cameras with the robot base and record the RGB-D streams from the cameras and robot joint pose simultaneously during the data collection. We train SAM-E in 5 tasks with 10 episodes for each, including \textit{put the towel on the cabinet}, \textit{stack the block}, \textit{close the drawer}, \textit{pick up the banana}, and \textit{put the orange into the drawer}. All the episodes are collected by human demonstrators. The results show that SAM-E can perform real-time prediction in real-world scenarios and complete tasks effectively, validating SAM-E's capability in real-world scenarios. See the \S\ref{app_realworld} and the videos for more details and model performance.

\begin{table}[t]
\centering
\caption{Ablation over action sequence length $h$}
\label{tab:ablation_horizon}
\resizebox{0.85\columnwidth}{!}{
\begin{tabular}{@{}lcc@{}}
\toprule
Models & Success Rate    & Inference Time (ms)    \\
\midrule
RVT           & 62.9            &  103\\
SAM-E~($h=1$)  & 30.6$\pm$1.4    &  126\\
SAM-E~($h=3$)  & 64.0$\pm$0.6    &  144\\
SAM-E~($h=5$)  & 70.6$\pm$0.7    &  152\\
SAM-E~($h=7$)  & 66.5$\pm$1.2    &  156\\
\bottomrule
\end{tabular}
}
\vspace{-1em}
\end{table}

\section{Conclusion}
We have introduced \textbf{S}egment \textbf{A}nything \textbf{M}odel for \textbf{E}mbodied 3D manipulation (SAM-E), a novel multi-view architecture that adopts SAM as the visual foundation model with parameter-efficient finetuning for promptable perception to embodied scenarios, as well as a novel action-sequence prediction head for efficient planning and coherent execution. We conduct experiments of SAM-E on various 3D instruction-following tasks from RLBench for multi-task experiments and few-show adaptation. We find SAM-E outperforms prior state-of-the-art models on multi-task manipulation and achieves a significant improvement in execution efficiency and few-shot adaptation with great generalization ability. Our work highlights the feasibility of leveraging a visual foundation model and sequence prediction for enhancing generalization and efficiency in 3D manipulation.

\section*{Acknowledgements}

This work was supported by the National Natural Science Foundation of China (Nos.62306242 \& 62376222), the STI 2030-Major Projects under Grant 2021ZD0201404, the National Key R\&D Program of China (No.2022ZD0160102), and Young Elite Scientists Sponsorship Program by CAST (No.2023QNRC001). We thank Wenke Xia for his excellent assistance in hardware deployment and data collection for real robot experiments.

\section*{Impact Statement}
This paper presents work whose goal is to advance the field of 
Machine Learning. There are many potential societal consequences 
of our work, none which we feel must be specifically highlighted here.

\nocite{langley00}

\bibliography{main}
\bibliographystyle{icml2024}

\newpage
\appendix
\onecolumn



\section{RLBench Tasks}
\label{app:task}

\begin{figure}[!ht]
\begin{center}
\centerline{\includegraphics[width=\columnwidth]{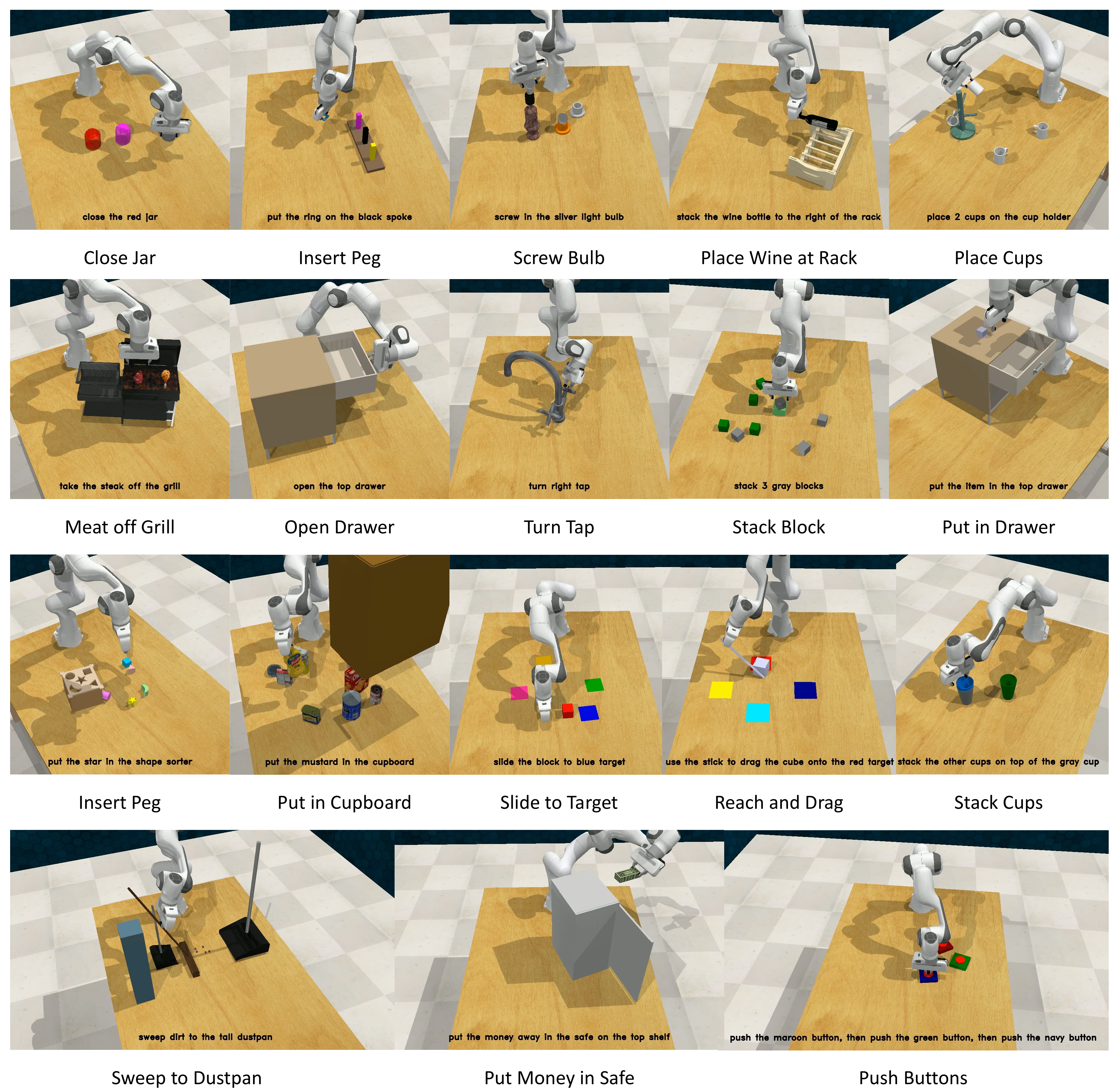}}
\caption{\textbf{Language-Conditioned Manipulation Tasks in RLBench}. We conduct multi-task experiments on 18 simulated tasks in RLBench\cite{rlbench}. Apart from the language instruction depicted in the figures, there are a total of 249 variations of these tasks, as illustrated in Table~\ref{appendix_tab:rlbench_task0}. During the test, the agent needs to handle the novel object poses, randomly sampled goals, and randomly sampled scenes with different semantic instantiations of object colors, shapes, sizes, and categories within a maximum of 25 execution steps.}
\label{appendix_fig:multi-task}
\end{center}
\vskip -0.2in
\end{figure}

\begin{figure}[!ht]
\begin{center}
\centerline{\includegraphics[width=\columnwidth]{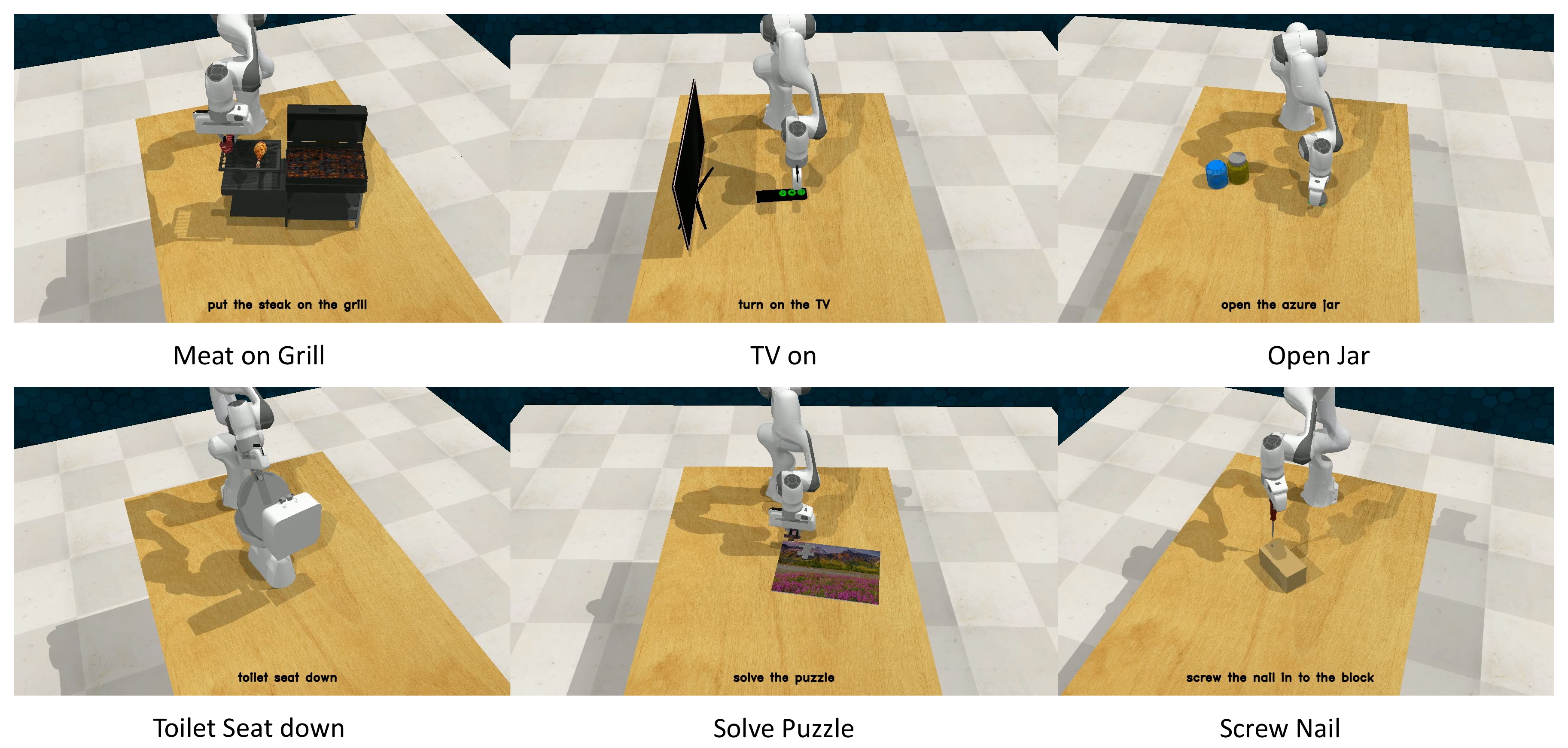}}
\caption{\textbf{Language-Conditioned Manipulation Tasks in RLBench}. We conduct few-shot adaptation experiments on 6 simulated tasks in RLBench to evaluate the generalization ability of SAM-E. Task variations are shown in Table~\ref{appendix_tab:rlbench_task1}. The tasks must be completed by the agent within a maximum of 25 steps.}
\label{appendix_fig:few-shot}
\end{center}
\vskip -0.2in
\end{figure}

We follow the multi-task multi-variation simulated experiments setting of RVT \cite{RVT} and PerAct \cite{peract} with 18 RLBench tasks (shown in Figure \ref{appendix_fig:multi-task}) and 249 unique variations across object placement, color, size, category, count, and shape. Here we give a summary of the 18 RLBench tasks in Table \ref{appendix_tab:rlbench_task0}. The extra 6 RLBench tasks (shown in Figure \ref{appendix_fig:few-shot}) for the few-shot adaptation experiment are summarized in Table \ref{appendix_tab:rlbench_task1}.

\begin{table}[!h]
\centering
\caption{The 18 RLBench tasks for multi-task experiment}
\vspace{0.2em}
\label{appendix_tab:rlbench_task0}
\resizebox{\columnwidth}{!}{
\begin{tabular}{@{}llccl@{}}
\toprule
Task name & Language Template & Avg. Keyframes & \#of Variations & Variation Type\\
\midrule     
put in drawer                   & “put the item in the \rule{0.3cm}{0.4pt} drawer”                                       &12.0  &3   &placement        \\
reach and drag                  & “use the stick to drag the cube onto the \rule{0.3cm}{0.4pt} target”                   &6.0   &20  &color        \\
turn tap                        & “turn \rule{0.3cm}{0.4pt} tap”                                                         &2.0   &2   &placement        \\
slide to target                 & “slide the block to \rule{0.3cm}{0.4pt} target”                                        &4.7   &4   &color        \\
open drawer                     & “open the \rule{0.3cm}{0.4pt} drawer”                                                  &3.0   &3   &placement        \\
put in cupboard                 & “put the \rule{0.3cm}{0.4pt} in the cupboard”                                          &5.0   &9   &category        \\
place in shape sorter           & “put the \rule{0.3cm}{0.4pt} in the shape sorter”                                      &5.0   &5   &shape        \\
put money in safe               & “put the money away in the safe on the \rule{0.3cm}{0.4pt} shelf”                      &5.0   &3   &placement        \\
push buttons                    & “push the \rule{0.3cm}{0.4pt} button, [then the \rule{0.3cm}{0.4pt} button]”           &3.8   &50  &color        \\
close jar                       & “close the \rule{0.3cm}{0.4pt} jar”                                                    &6.0   &20  &color         \\
stack block                     &  “stack \rule{0.3cm}{0.4pt} \rule{0.3cm}{0.4pt} blocks”                                &14.6  &60  &color,count  \\
place cups                      & “place \rule{0.3cm}{0.4pt} cups on the cup holder”                                     &11.5  &3   &count        \\
place wine at rack              & “stack the wine bottle to the \rule{0.3cm}{0.4pt} of the rack”                         &5.0   &3   &placement    \\
screw bulb                      & “screw in the \rule{0.3cm}{0.4pt} light bulb”                                          &7.0   &20  &color        \\
sweep to dustpan                &“sweep dirt to the \rule{0.3cm}{0.4pt} dustpan”                                         &4.6   &2   &size         \\
insert peg                      &“put the ring on the \rule{0.3cm}{0.4pt} spoke”                                         &5.0   &20  &color        \\
meat off grill                  &“take the \rule{0.3cm}{0.4pt} off the grill”                                            &5.0   &2   &category     \\
stack cups                      &“stack the other cups on top of the \rule{0.3cm}{0.4pt} cup”                            &10.0  &20  &color        \\
\bottomrule
\end{tabular}
}
\end{table}

\begin{table}[!h]
\centering
\caption{The 6 RLBench tasks used for the few-shot adaptation experiments.}
\label{appendix_tab:rlbench_task1}
\begin{tabular}{@{}llccl@{}}
\toprule
Task name & Language Template & Avg. Keyframes & \#of Variations & Variation Type\\
\midrule     
meat on grill        & “put the \rule{0.3cm}{0.4pt} on the grill ”                     &5.0  &2            &category \\
open jar             & “open the \rule{0.3cm}{0.4pt} jar”                              &6.0  &20           &color    \\
screw nail           & “screw the nail in to the block”                                &6.0  &1            &-        \\
toilet seat down     & “toilet seat down”                                              &4.7  &1            &-        \\
tv on                & “turn on the TV”                                                &8.0  &1            &-        \\
solve puzzle         & “solve the puzzle”                                              &5.0  &1            &-        \\
\bottomrule
\end{tabular}
\end{table}

\section{Implementation Details}
\label{app:implement}
In this section, we provide more implementation details of SAM-E.
\subsection{Visual Input}
In our experiments of RLBench, the visual observations are captured by four cameras (left shoulder, right shoulder, front, and wrist) with a resolution of 128 $\times$ 128 in RGB-D. We follow the re-render approach introduced by RVT \cite{RVT} before feeding visual images to the model. Specifically, the RGB-D images are 
rerendered to generate virtual images in the form of cube orthographic projection. Then we use the cube orthographic projections as the visual inputs of SAM-E.


\subsection{Action Sequence Imitation}
\label{app:horizon}
We utilize a multi-channel action sequence policy head to predict the action sequence, trained by action sequence imitation. To extract the temporal information of actions from the expert demonstrations, we employ the keyframe extraction on each demonstration, generating a dataset of keyframe sequences. Given observations, SAM-E generates an action sequence with a default action horizon of 5 and is trained to maximize the likelihood objective of imitation learning. Note that the action sequence data may have variable lengths, when the data is shorter than the action horizon, we mask the untrained action head, and when the data is longer, we truncate it accordingly.

\subsection{Hyperparameters}
In our experiments, the hyperparameters are primarily fixed, as shown in Table~\ref{appendix_tab:hyperparameter}.
\begin{table}[!htbp]
\centering
\caption{Training Hyperparameters}
\vspace{0.2em}
\label{appendix_tab:hyperparameter}
\begin{tabular}{@{}cccc@{}}
\toprule
Hyperparameters & Multi-task Training   & Few-shot adaptation   \\
\midrule     
batch size        &10                &10             \\
learning rate        &4e-3           &4e-3             \\
optimizer          &LAMB            &LAMB               \\
learning rate schedule  &cosine decay &cosine decay     \\
warmup steps        &2000           &2000               \\
training steps     & 60K             & 4K                  \\
training epochs     &15             &1                  \\
\bottomrule
\end{tabular}
\end{table}

\section{Visualization}
\label{app:visualization}
We visualize the attention map of the multi-view transformer to show SAM-E's various attention patterns for task comprehension and action sequence prediction. We use task \textit{put\_item\_in\_drawer} as an example, which is completed by three executions. 

\begin{figure}[!htbp]
\vskip -0.1in
\begin{center}
\centerline{\includegraphics[width=\columnwidth]{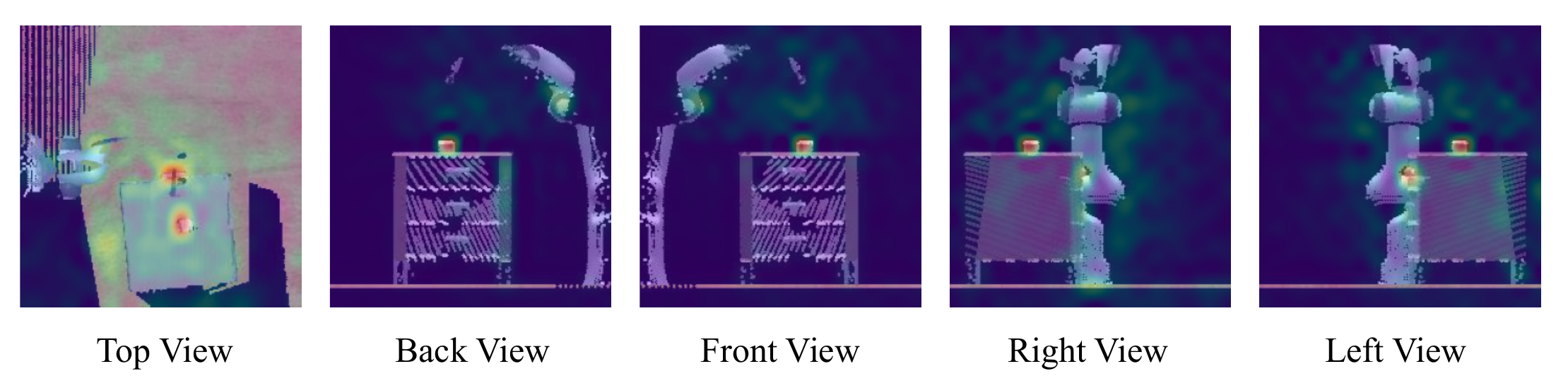}}
\caption{SAM-E's multi-view attention map of the initial inference.}
\label{appendix_fig:app_attn_0}
\end{center}
\vskip -0.2in
\end{figure}

(\romannumeral1) In the first execution with the initial observation (see Figure~\ref{appendix_fig:app_attn_0}), SAM-E's attention, from one of its heads, is predominantly focused on the Franka robot, the drawer cabinet, and more specifically, the item on the cabinet and the handle of the top drawer. This observation aligns with the given instruction to `\emph{put the item in the top drawer}', highlighting SAM-E's capability to identify key objects within the scene according to the task description for task execution.

\begin{figure}[!htbp]
\vskip -0.1in
\begin{center}
\centerline{\includegraphics[width=\columnwidth]{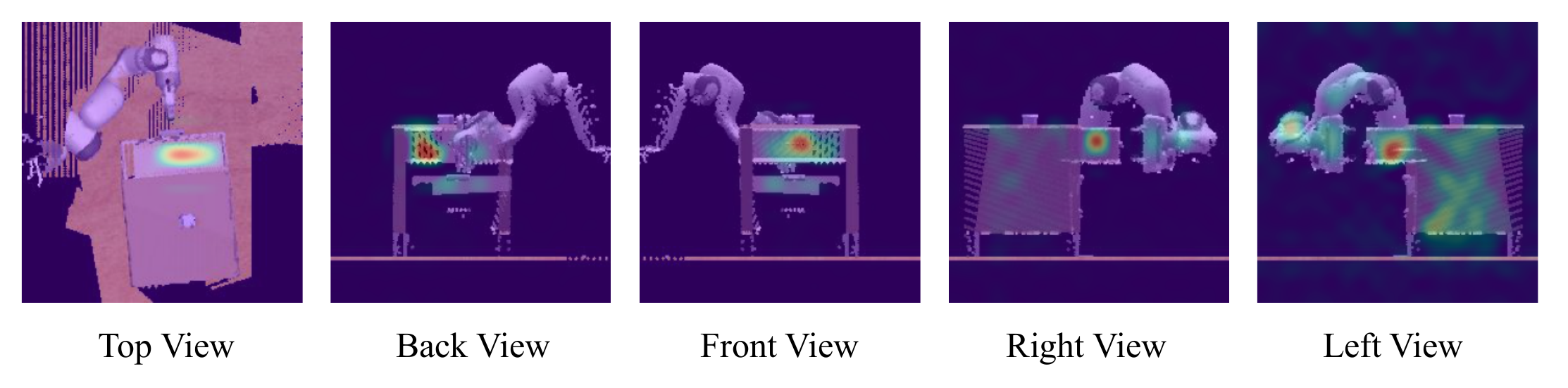}}
\caption{SAM-E's multi-view attention map of the second inference, focusing on the open drawer.}
\label{appendix_fig:app_attn_1_0}
\end{center}
\vskip -0.2in
\end{figure}

\begin{figure}[!htbp]
\vskip -0.1in
\begin{center}
\centerline{\includegraphics[width=\columnwidth]{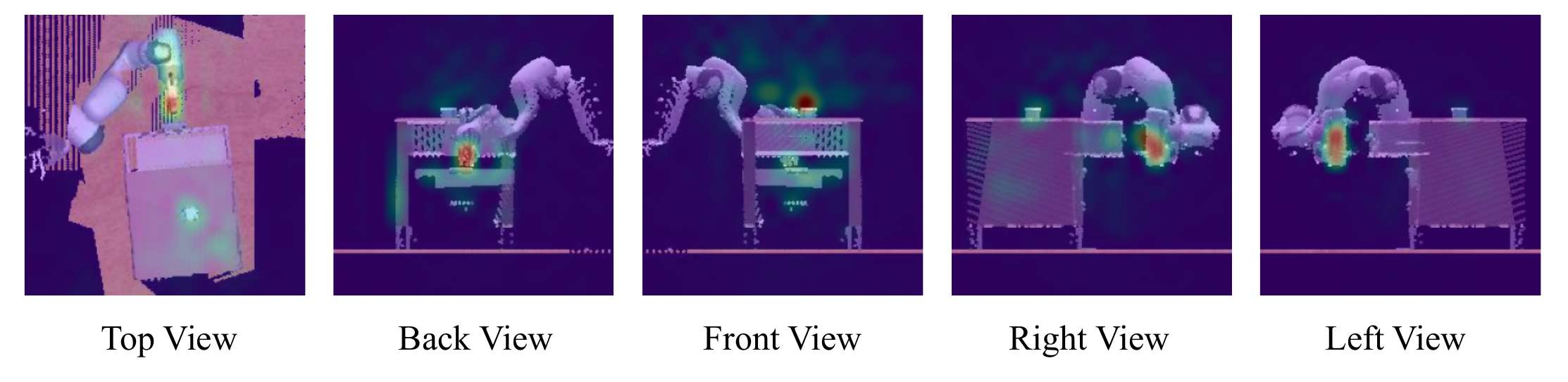}}
\caption{SAM-E's multi-view attention map of the second inference, focusing on the end-effector and the item.}
\label{appendix_fig:app_attn_1_1}
\end{center}
\vskip -0.2in
\end{figure}

(\romannumeral2) In the second inference, following an action sequence that results in the opening of the top drawer, SAM-E adapts its focus. It now observes the newly available space within the drawer for placing the item (see Figure~\ref{appendix_fig:app_attn_1_0}). Concurrently, another of its attention heads redirects back to the end-effector and the item, strategizing the subsequent action of picking up and placing the item into the drawer(see Figure~\ref{appendix_fig:app_attn_1_1}).

\begin{figure}[!htbp]
\vskip -0.1in
\begin{center}
\centerline{\includegraphics[width=\columnwidth]{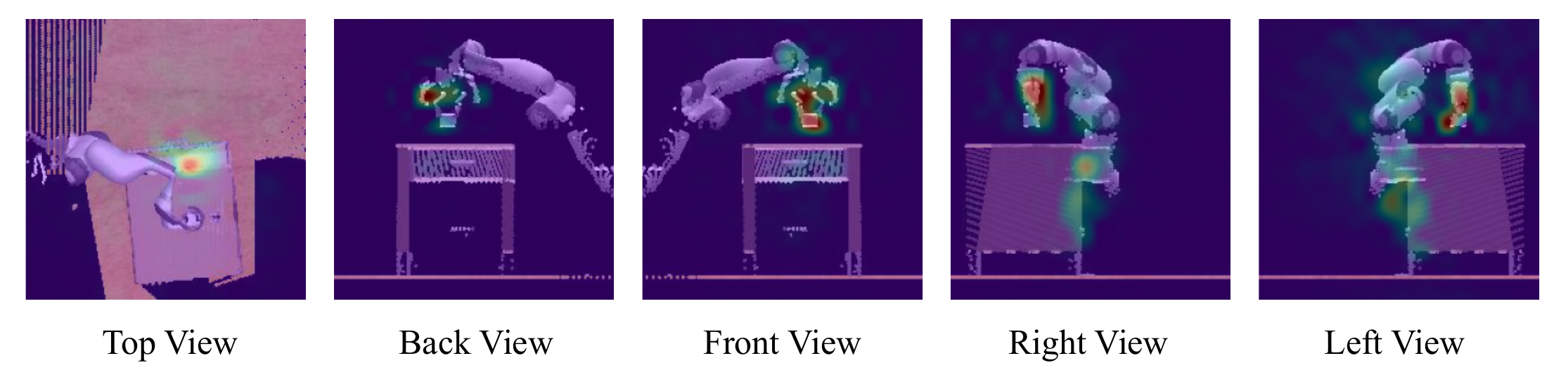}}
\caption{SAM-E's multi-view attention map of the last inference.}
\label{appendix_fig:app_attn_2}
\end{center}
\vskip -0.2in
\end{figure}

(\romannumeral3) In the final inference (see Figure~\ref{appendix_fig:app_attn_2}), SAM-E concentrates on the end-effector picking up the item and positioning it accurately into the target position. This phase likely involves precise adjustments and movements, ensuring the successful completion of the task as the language instruction.

\newpage
\section{One Glance Results}
\label{app:infer}

\begin{figure}[!htbp]
\begin{center}
\centerline{\includegraphics[width=0.8\columnwidth]{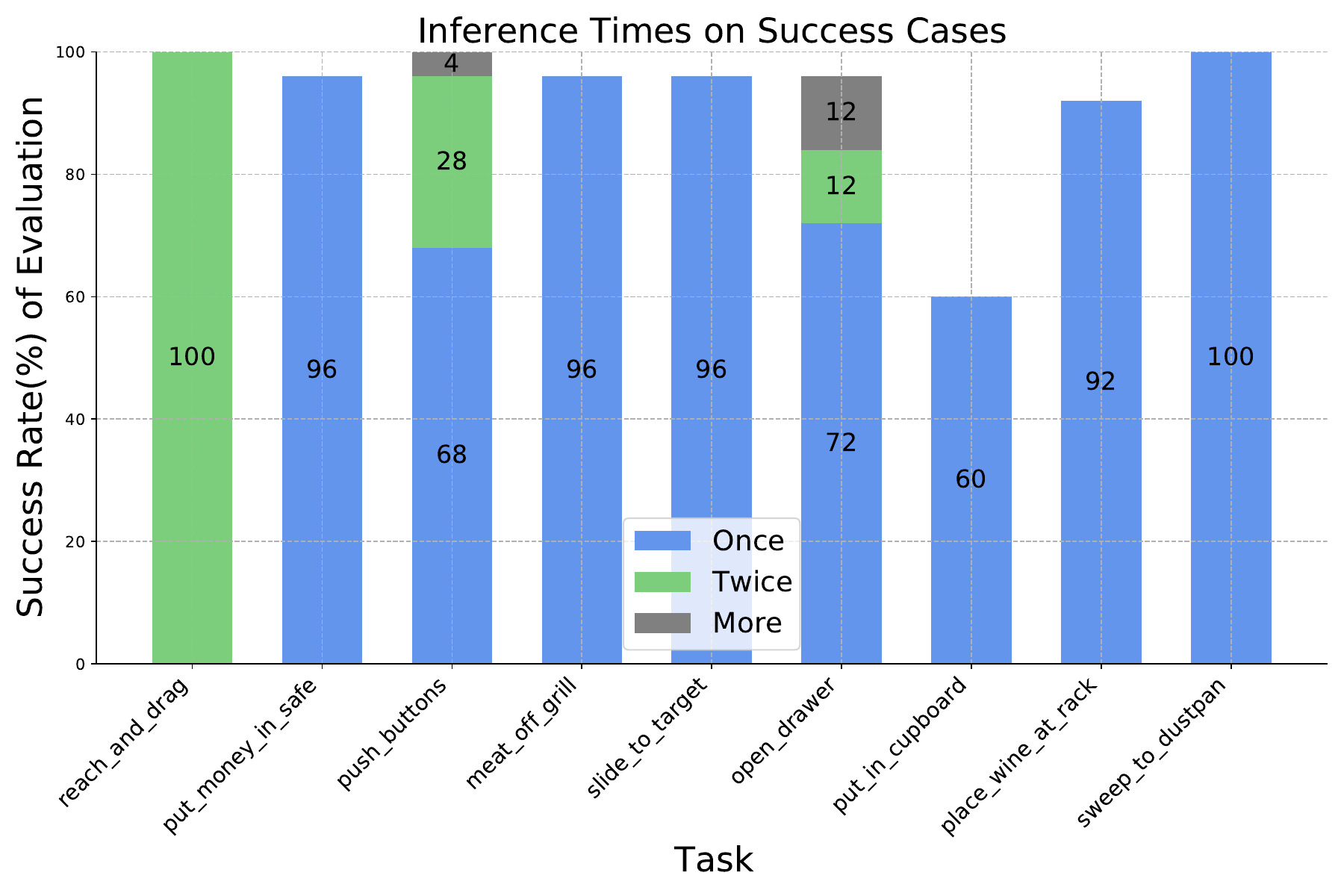}}
\vspace{-1em}
\caption{The comparison of execution times on success cases of several tasks.}
\label{appendix_fig:one_glance}
\end{center}
\vskip -0.2in
\end{figure}

Figure~\ref{appendix_fig:one_glance} shows the results of execution times of SAM-E on success cases of several tasks. Thanks to its promptable perception and efficient action sequence prediction, SAM-E excels in task completion by executing actions coherently, resulting in improved performance and significantly reduced inference requirements. For the following tasks, in comparison, RVT requires an average of $[6.4, 6.0, 3.8, 5.5, 3.7, 4.3, 5.5, 5.0, 4.8]$ execution times of its success cases.

\section{Comparison with Hiveformer}
\label{app_hive}
To compare the performance of SAM-E and HiverFormer, we add experiments to train SAM-E with the same 10 tasks evaluated in the Hiveformer paper with 100 demonstrations per task, which is the same as Hiveformer (results are shown in Table \ref{appendix_tab:hivformer}). The score of Hiveformer is adapted from their original paper. We remark that SAM-E is trained with 10 tasks with \textbf{all variations}, which is much more challenging than Hiveformer which is trained with a unique variation for each task.

\begin{table*}[h]
\centering
\caption{\textbf{Comparison with Hiveformer.} Scores of Hiveformer are adopted from \citet{hiveFormer}. Mean and std of 5 evaluations are reported.}
\label{appendix_tab:hivformer}
\resizebox{\textwidth}{!}{
\begin{tabular}{@{}lccccccccccc@{}}
\toprule
Models  & \begin{tabular}[c]{@{}c@{}}Pick and\\ Lift\end{tabular} & \begin{tabular}[c]{@{}c@{}}Pick up\\ Cup\end{tabular} & \begin{tabular}[c]{@{}c@{}}Put Knife\\ on Board\end{tabular} & \begin{tabular}[c]{@{}c@{}}Reach\\ Target\end{tabular} & \begin{tabular}[c]{@{}c@{}}Stack \\ Wine\end{tabular} & \begin{tabular}[c]{@{}c@{}}Take Money \\ out Safe\end{tabular} 
&\begin{tabular}[c]{@{}c@{}}Take Umbrella\\ out Stand\end{tabular} 
& \begin{tabular}[c]{@{}c@{}}Push\\ Buttons\end{tabular} 
& \begin{tabular}[c]{@{}c@{}}Put Money\\ in Safe\end{tabular}
& \begin{tabular}[c]{@{}c@{}}Slide to\\ Target\end{tabular}

&\multicolumn{1}{|c}{\begin{tabular}[c]{@{}c@{}}\textbf{On}\\ \textbf{Average}\end{tabular}} \\
\midrule
Hiveformer &88.9	&92.9	&75.3	&100.0	&71.2  &79.1	&89.2	&100.0	&58.2  &78.7  &\multicolumn{1}{|c}{83.3} \\
SAM-E  & 87.2$\pm$1.8 & 88.8$\pm$5.2 & 68.0$\pm$4.0 & 100.0$\pm$0.0 & 69.6$\pm$7.3 & 98.4$\pm$2.2 & 96.0$\pm$2.8 & 100.0$\pm$0.0 &93.6$\pm$2.2  &84.8$\pm$7.7  & \multicolumn{1}{|c}{\cellcolor{blue!20}\textbf{88.6}$\pm$0.7} \\
\bottomrule
\end{tabular}
}
\vspace{-1em}
\end{table*}

\section{Ablation}
\label{app:ablation}
We provide the complete results of the ablation study in Table \ref{appendix_tab:ablation} and Table \ref{appendix_tab:ablation_horizon}.

\begin{table*}[!htbp]
\centering
\caption{\textbf{Ablation Performances of SAM-E's variations.} Mean and std of 5 evaluations are reported.}
\vspace{0.2em}
\label{appendix_tab:ablation}
\resizebox{\textwidth}{!}{
\begin{tabular}{@{}lcccccccccc@{}}
\toprule
Models & \begin{tabular}[c]{@{}c@{}}Put in\\ Drawer\end{tabular} & \begin{tabular}[c]{@{}c@{}}Reach \\and Drag\end{tabular} & \begin{tabular}[c]{@{}c@{}}Turn\\ Tap\end{tabular} & \begin{tabular}[c]{@{}c@{}}Slide to\\ Target\end{tabular} & \begin{tabular}[c]{@{}c@{}}Open\\ Drawer\end{tabular} & \begin{tabular}[c]{@{}c@{}}Put in\\ Cupboard\end{tabular} & \begin{tabular}[c]{@{}c@{}}Place in\\ Shape Sorter\end{tabular} & \begin{tabular}[c]{@{}c@{}}Put Money\\ in Safe\end{tabular} & \begin{tabular}[c]{@{}c@{}}Push\\ Buttons\end{tabular} & \begin{tabular}[c]{@{}c@{}}Close\\ Jar\end{tabular}\\ 
\midrule
SAM-E & 92.0$\pm$5.7 & \textbf{100.0}$\pm$0.0 & \textbf{100.0}$\pm$0.0 & \textbf{95.2}$\pm$1.8 & \textbf{95.2}$\pm$5.2 & \textbf{64.0}$\pm$2.8 & 34.4$\pm$6.1 & \textbf{95.2}$\pm$3.3 & \textbf{100.0}$\pm$0.0 & 82.4$\pm$3.6 \\
SAM-E~(LoRA, QKV)  & 88.8$\pm$7.7	&\textbf{100.0}$\pm$0.0	&\textbf{100.0}$\pm$0.0	&90.4$\pm$6.1	&94.4$\pm$3.6	&57.6$\pm$4.6	&\textbf{37.6}$\pm$5.4	&92.8$\pm$5.2	&\textbf{100.0}$\pm$0.0	&78.4$\pm$4.6\\
SAM-E~(w/o LoRA) & 84.8$\pm$7.2	&\textbf{100.0}$\pm$0.0	&\textbf{100.0}$\pm$0.0	&92.0$\pm$4.0	&92.8$\pm$3.3	&52.0$\pm$2.8	&31.2$\pm$5.2	&92.8$\pm$1.8	&98.4$\pm$2.2	&\textbf{87.2}$\pm$5.2\\
SAM-E~(full finetune) & \textbf{93.3}$\pm$4.6	&98.7$\pm$2.3	&\textbf{100.0}$\pm$0.0	&69.3$\pm$11.5	&90.7$\pm$2.3	&52.0$\pm$0.0	&29.3$\pm$11.5	&89.3$\pm$2.3	&\textbf{100.0}$\pm$0.0	&68.0$\pm$0.0\\
\bottomrule

Models  & \begin{tabular}[c]{@{}c@{}}Stack\\ Blocks\end{tabular} & \begin{tabular}[c]{@{}c@{}}Place\\ Cups\end{tabular} & \begin{tabular}[c]{@{}c@{}}Place Wine\\ at Rack\end{tabular} & \begin{tabular}[c]{@{}c@{}}Screw\\ Bulb\end{tabular} & \begin{tabular}[c]{@{}c@{}}Sweep to\\ Dustpan\end{tabular} & \begin{tabular}[c]{@{}c@{}}Insert \\ Peg\end{tabular} &
\begin{tabular}[c]{@{}c@{}}Meat off\\ Grill\end{tabular} & 
\begin{tabular}[c]{@{}c@{}}Stack\\ Cups\end{tabular} &
\multicolumn{1}{|c}{\begin{tabular}[c]{@{}c@{}}\textbf{On}\\ \textbf{Average}\end{tabular}} &
\begin{tabular}[c]{@{}c@{}}\textbf{Inference}\\ \textbf{Steps(Sum)}\end{tabular}\\
\midrule
SAM-E  & 26.4$\pm$4.6 & 0.0$\pm$0.0 & \textbf{94.4}$\pm$4.6 & \textbf{78.4}$\pm$3.6 & \textbf{100.0}$\pm$0.0 & \textbf{18.4}$\pm$4.6 & 95.2$\pm$3.3 & 0.0$\pm$0.0 & \multicolumn{1}{|c}{\cellcolor{blue!20}\textbf{70.6}$\pm$0.7} &\cellcolor{blue!20}\textbf{1130}$\pm$12\\
SAM-E~(LoRA, QKV)    &\textbf{32.0}$\pm$4.9	&\textbf{3.2}$\pm$1.8	&93.6$\pm$3.6	&69.6$\pm$4.6	&98.4$\pm$2.2	&6.4$\pm$6.1	&97.6$\pm$2.2	&5.6$\pm$2.2	&\multicolumn{1}{|c}{69.2$\pm$0.9}	&1142$\pm$6 \\
SAM-E~(w/o LoRA)   &20.8$\pm$7.2	&0.0$\pm$0.0	   &92.0$\pm$4.0	     &64.0$\pm$7.5    &96.8$\pm$1.8	&8.8$\pm$6.6	&94.4$\pm$3.6	&0.8$\pm$1.8	&\multicolumn{1}{|c}{67.2$\pm$1.0} &1182$\pm$6 \\
SAM-E~(full finetune) &20.0$\pm$6.9	&1.3$\pm$2.3	&90.7$\pm$2.3	&69.3$\pm$4.6	&\textbf{100.0}$\pm$0.0	      &0.0$\pm$0.0	   &\textbf{98.7}$\pm$2.3	&\textbf{13.3}$\pm$4.6	&\multicolumn{1}{|c}{65.8$\pm$1.0} &1204$\pm$18 \\

\bottomrule
\end{tabular}
}
\end{table*}

\begin{table*}[!htbp]
\centering
\caption{\textbf{Ablation Performances with different action sequence length.} Mean and std of 5 evaluations are reported.}
\vspace{0.2em}
\label{appendix_tab:ablation_horizon}
\resizebox{\textwidth}{!}{
\begin{tabular}{@{}lcccccccccc@{}}
\toprule
Models & \begin{tabular}[c]{@{}c@{}}Put in\\ Drawer\end{tabular} & \begin{tabular}[c]{@{}c@{}}Reach \\and Drag\end{tabular} & \begin{tabular}[c]{@{}c@{}}Turn\\ Tap\end{tabular} & \begin{tabular}[c]{@{}c@{}}Slide to\\ Target\end{tabular} & \begin{tabular}[c]{@{}c@{}}Open\\ Drawer\end{tabular} & \begin{tabular}[c]{@{}c@{}}Put in\\ Cupboard\end{tabular} & \begin{tabular}[c]{@{}c@{}}Place in\\ Shape Sorter\end{tabular} & \begin{tabular}[c]{@{}c@{}}Put Money\\ in Safe\end{tabular} & \begin{tabular}[c]{@{}c@{}}Push\\ Buttons\end{tabular} & \begin{tabular}[c]{@{}c@{}}Close\\ Jar\end{tabular}\\ 
\midrule
SAM-E~($h=1$) & 0.0$\pm$0.0 & 6.7$\pm$2.3 & 98.7$\pm$2.3 & 45.3$\pm$4.6 & 72.0$\pm$6.9 & 8.0$\pm$4.0 & 14.7$\pm$2.3 & 8.0$\pm$0.0 & 69.3$\pm$2.3 & 12.0$\pm$4.0 \\
SAM-E~($h=3$) & 77.6$\pm$2.2 & 84.8$\pm$1.8 & \textbf{100.0}$\pm$0.0 & 72.8$\pm$1.8 & 92.0$\pm$2.8 & 31.2$\pm$5.2 & \textbf{35.2}$\pm$3.3 & 84.8$\pm$5.2 & 99.2$\pm$1.8 & 73.6$\pm$3.6 \\
SAM-E~($h=5$) & \textbf{92.0}$\pm$5.7 & \textbf{100.0}$\pm$0.0 & \textbf{100.0}$\pm$0.0 & \textbf{95.2}$\pm$1.8 & \textbf{95.2}$\pm$5.2 & \textbf{64.0}$\pm$2.8 & 34.4$\pm$6.1 & \textbf{95.2}$\pm$3.3 & \textbf{100.0}$\pm$0.0 & \textbf{82.4}$\pm$3.6 \\
SAM-E~($h=7$) & 88.8$\pm$9.5 & 99.2$\pm$1.8 & \textbf{100.0}$\pm$0.0 & 80.8$\pm$18.6 & 90.4$\pm$4.6 & 53.6$\pm$7.3 & 28.8$\pm$3.3 & 92.8$\pm$1.8 & \textbf{100.0}$\pm$0.0 & 72.8$\pm$3.3 \\
\bottomrule

Models  & \begin{tabular}[c]{@{}c@{}}Stack\\ Blocks\end{tabular} & \begin{tabular}[c]{@{}c@{}}Place\\ Cups\end{tabular} & \begin{tabular}[c]{@{}c@{}}Place Wine\\ at Rack\end{tabular} & \begin{tabular}[c]{@{}c@{}}Screw\\ Bulb\end{tabular} & \begin{tabular}[c]{@{}c@{}}Sweep to\\ Dustpan\end{tabular} & \begin{tabular}[c]{@{}c@{}}Insert \\ Peg\end{tabular} &
\begin{tabular}[c]{@{}c@{}}Meat off\\ Grill\end{tabular} & 
\begin{tabular}[c]{@{}c@{}}Stack\\ Cups\end{tabular} &
\multicolumn{1}{|c}{\begin{tabular}[c]{@{}c@{}}\textbf{On}\\ \textbf{Average}\end{tabular}} &
\begin{tabular}[c]{@{}c@{}}\textbf{Inference}\\ \textbf{Steps(Sum)}\end{tabular}\\
\midrule
SAM-E~($h=1$)  & 0.0$\pm$0.0 & 1.3$\pm$2.3 & 40.0$\pm$6.9 & 58.7$\pm$2.3 & 24.0$\pm$4.0 & 34.7$\pm$14.0 & 54.7$\pm$4.6 & 2.7$\pm$4.6 & \multicolumn{1}{|c}{30.6$\pm$1.4} & 8329$\pm$60\\
SAM-E~($h=3$)     &16.8$\pm$3.3	&1.6$\pm$2.2	&76.8$\pm$4.4   &49.6$\pm$6.7	&87.2$\pm$1.8	&\textbf{54.4}$\pm$2.2	&\textbf{100.0}$\pm$0.0	&\textbf{13.6}$\pm$3.6	&\multicolumn{1}{|c}{64.0$\pm$0.6}	&2026$\pm$30 \\
SAM-E~($h=5$)    & \textbf{26.4}$\pm$4.6 & 0.0$\pm$0.0 & \textbf{94.4}$\pm$4.6 & \textbf{78.4}$\pm$3.6 & \textbf{100.0}$\pm$0.0 & 18.4$\pm$4.6 & 95.2$\pm$3.3 & 0.0$\pm$0.0 & \multicolumn{1}{|c}{\cellcolor{blue!20}\textbf{70.6}$\pm$0.7} &1130$\pm$12\\
SAM-E~($h=7$)  &13.6$\pm$5.4	&\textbf{3.2}$\pm$3.3	&92.0$\pm$4.0	&70.4$\pm$5.4	&\textbf{100.0}$\pm$0.0	&8.8$\pm$4.4	   &97.6$\pm$3.6	&4.8$\pm$1.8	&\multicolumn{1}{|c}{66.5$\pm$1.2} &\cellcolor{blue!20}\textbf{919}$\pm$12 \\

\bottomrule
\end{tabular}
}
\end{table*}

\section{Real-World Experiments}
\label{app_realworld}

\begin{figure}[!htbp]
\begin{center}
\centerline{\includegraphics[width=\columnwidth]{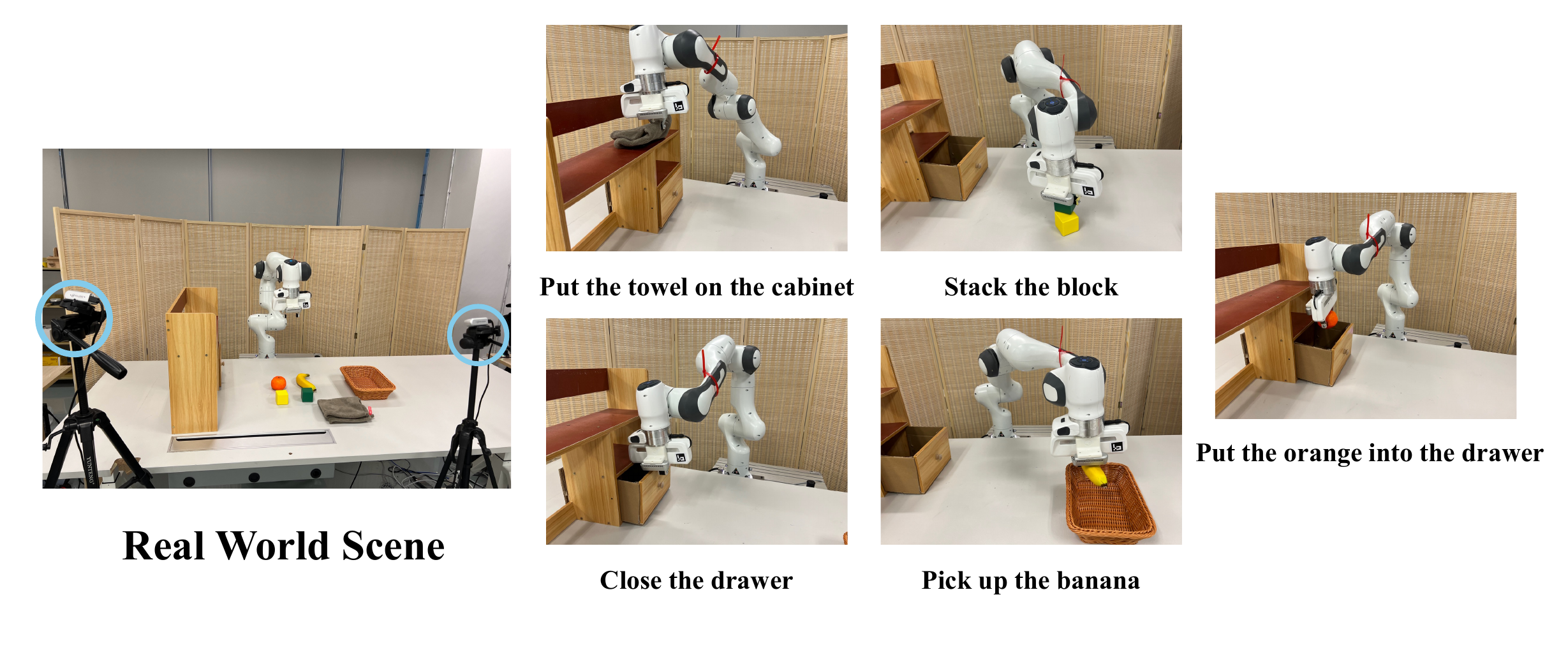}}
\vspace{-1em}
\caption{Real-World Scene and tasks.}
\label{appendix_fig:realworld}
\end{center}
\end{figure}

We conduct real-world experiments on a FranKa Panda robot arm in the real world, equipped with a dual RGB-D camera setup positioned at the left front and right front for multi-view observation, shown in Figure \ref{appendix_fig:realworld}. We construct the real-world scene and design 5 tasks for experiments, including \textit{put the towel on the cabinet}, \textit{stack the block}, \textit{close the drawer}, \textit{pick up the banana}, and \textit{put the orange into the drawer}. For data collection, we manually control the robot arm for demonstrations by a controller and collect the RGB-D stream and robot joint pose simultaneously with a data collection pipeline. We collect demonstrations with variations in item placement for all tasks. See \url{https://sam-embodied.github.io/} for videos and performance. 

\section{Limitation and Future Work}
In this work, we propose SAM-E with SAM as the visual foundation and action-sequence policy head, which outperforms prior state-of-the-art methods. However, we also identify limitations that suggest directions for future research. We employ parameter-efficient fine-tuning on relatively limited robot data to enhance its understanding of embodied manipulation. Future improvements might include leveraging the scalability of the visual foundation through training on larger datasets, such as Open-X \cite{Open-X}. Additionally, we employed a fixed horizon for the action-sequence policy, which, while generally effective, could be less suitable for certain tasks, such as \textit{stack cups} in our experiments, in which may need to pay more attention to the trade-off between precision and coherence of the action. It would be intriguing to see the action horizon optimized through a mechanism or learned from data.

\end{document}